\documentclass[lettersize, journal]{IEEEtran}
\usepackage{amsmath}
\usepackage{mleftright}
\usepackage{amssymb}
\usepackage{mathtools}
\usepackage{bm}
\usepackage{xcolor}
\usepackage{float}
\usepackage{fancyvrb}
\usepackage{hyperref}
\usepackage{graphicx}
\usepackage{lipsum}
\usepackage{algorithm}
\usepackage{algpseudocode}
\usepackage{todonotes}
\usepackage{booktabs}
\usepackage{makecell}
\usepackage{enumitem}
\usepackage{adjustbox}

\usepackage{pifont}
\newcommand{\cmark}{\ding{51}}
\newcommand{\xmark}{\ding{55}}

\usepackage{subcaption}
\usepackage[noabbrev,capitalize,nameinlink]{cleveref}
\creflabelformat{equation}{#2\textup{#1}#3}
\usepackage{array}
\usepackage{adjustbox}
\newcommand*\rot{\rotatebox[origin=l]{90}}

\floatstyle{plaintop}
\restylefloat{table}

\usepackage{siunitx}

\usepackage{multirow}

\newcommand{\LineComment}[2][0]{\Statex\ifnum#1>0 \hspace{\dimexpr #1\dimexpr \algorithmicindent \relax \relax} \fi\(\triangleright\) #2}
\let\oldnl\nl 
\newcommand{\nonl}{\renewcommand{\nl}{\let\nl\oldnl} }

\usepackage{tikz}
\usetikzlibrary{shadows}

\usepackage{soulutf8}
\sethlcolor{yellow}

\newcommand{\indep}{\perp \!\!\! \perp}

\usepackage[most]{tcolorbox}
\usepackage{varwidth}
\definecolor{template-gray}{rgb}{0.6, 0.6, 0.6}

\tcbset{%
  boxenv/.style={
    enhanced, breakable,
    before skip=2mm, after skip=2mm,
    colback=template-gray!5,
	colframe=template-gray!100,
    attach boxed title to top left={xshift=1cm, yshift*=1mm-\tcboxedtitleheight},
    varwidth boxed title*=-3cm,
    boxed title style={frame code={
        \path[fill=tcbcolback!30!black]
          ([yshift=-1mm,xshift=-1mm]frame.north west)
          arc[start angle=0,end angle=180,radius=1mm]
          ([yshift=-1mm,xshift=1mm]frame.north east)
          arc[start angle=180,end angle=0,radius=1mm];
        \path[left color=tcbcolback, right color=tcbcolback,
            middle color=tcbcolback]
          ([xshift=-2mm]frame.north west) -- ([xshift=2mm]frame.north east)
          [rounded corners=1mm]-- ([xshift=1mm,yshift=-1mm]frame.north east)
          -- (frame.south east) -- (frame.south west)
          -- ([xshift=-1mm,yshift=-1mm]frame.north west)
          [sharp corners]-- cycle;
        },
      interior engine=empty},
    colbacktitle=template-gray!100,
    fonttitle=\bfseries
  }
}
\newtcolorbox{boxenv}[1][]{boxenv,#1}

\usepackage[mincitenames=1,maxcitenames=1,style=ieee]{biblatex}
\usepackage[section]{placeins}

\begin{document}
\title{On the Illusion of Gender Bias in Face Recognition: Explaining the Fairness Issue Through Non-demographic Attributes}

\author{Paul Jonas Kurz, Haiyu Wu, Rouqaiah Al-Refai, Kevin W. Bowyer, Philipp Terh\"orst
\vspace{-1.0em}
	\IEEEcompsocitemizethanks{
		\IEEEcompsocthanksitem Paul Jonas Kurz is with Paderborn University, Paderborn, Germany and Technical University of Darmstadt, Darmstadt, Germany.
		\IEEEcompsocthanksitem Haiyu Wu and Kevin W. Bowyer are with the University of Notre Dame, Indiana, USA.
        \IEEEcompsocthanksitem Philipp Terh\"orst is with Johannes Gutenberg University Mainz, Mainz, Germany.
        \IEEEcompsocthanksitem Rouqaiah Al-Refai is with Paderborn University, Paderborn, Germany. Rouqaiah Al-Refai is the corresponding author.
	}
}

\markboth{Journal of \LaTeX\ Class Files,~Vol.~14, No.~8, August~2015}{Kurz \MakeLowercase{\textit{et al.}}:  On the "Illusion" of Gender Bias in Face Recognition: Explaining the Fairness Issue Through Non-demographic Attributes}

\vspace{-10mm}
\IEEEtitleabstractindextext{%
\begin{abstract}
Face recognition systems (FRS) exhibit significant accuracy differences based on the user’s gender. Since such a gender gap reduces the trustworthiness of FRS, more recent efforts have tried to find the causes. However, these studies make use of manually selected, correlated, and small-sized sets of facial features to support their claims. In this work, we analyze gender bias in face recognition by successfully extending the search domain to decorrelated combinations of 40 non-demographic facial characteristics. First, we introduce a toolchain to effectively decorrelate and aggregate facial attributes to enable a less-biased gender analysis on large-scale data. Second, we tailor two specialized metrics to quantify the effect of facial attributes on absolute and relative fairness. Based on these grounds, we thirdly present a novel unsupervised joint investigation framework capable of identifying attribute combinations leading to vanishing bias when used as filter predicates for balanced testing datasets. Experiments show the gender gap vanishing when images of male and female subjects share specific attributes, clearly indicating that the disparate performance is not a question of biology but of the social definition of appearance. These findings could reshape our understanding of fairness in face biometrics and provide insights into FRS, helping to address gender bias issues.

\end{abstract}

}

\maketitle

\IEEEdisplaynontitleabstractindextext

\IEEEpeerreviewmaketitle

\vspace{7mm}

\IEEEraisesectionheading{\section{Introduction}\label{sec:introduction}}

\IEEEPARstart{F}{ace} recognition systems (FRS) have been criticized as ``biased'', ``sexist'', or ``racist''~\cite{orcuttRaceBias, lohrRaceBias, hogginsRaceGenderBias, CastelvecchiCriticize, santowCriticize}.
Such criticisms often come in response to research work reporting that face recognition accuracy is lower for one demographic group than for another 
\cite{klareFaceRecognitionPerformance2012, race-phillipsRaceEffect, grotherFaceRecognitionVendor2019}. 
For instance, it is observed that female faces are much more likely to produce incorrect matching results than male faces.
This phenomenon is known as the ``gender gap'' \cite{bhattaGenderDifferenceHair2023}.
Since FRS are spreading worldwide, have a growing effect on daily life, and are increasingly used in critical decision-making processes, such as in forensics and law enforcement \cite{terhorstComprehensiveStudyFace2022, terhorstComparisonLevelMitigationEthnic2020},
these systems have a strong potential to discriminate against people on a wider scale.

As FRS are based on data-driven deep learning techniques, the initial speculation is the underrepresentation of female cohorts in the training dataset as a cause for inequality, broadly known as the gender gap ~\cite{albieroAnalysisGenderInequality2020}.
However,~\cite{albieroHowDoesGender2020} have shown that explicitly balancing the training data for number of male and female identities and images does not result in gender-balanced accuracy in the test data.
This motivates investigating the effect of non-demographic attributes~\cite{terhorstComprehensiveStudyFace2022}, such as facial hairstyle~\cite{wuFacialHairAttributeInDemographic2023, wuFacialHairPositionEffect2024}, makeup~\cite{albieroGenderedDifferences2022}, scalp hairstyle~\cite{bhattaGenderDifferenceHair2023}, face exposure~\cite{wuBrightnessInDemographic2023}, face morphology~\cite{albieroFaceRecognitionSexist2020a} as non-demographic attributes are strongly encoded in face representations \cite{DBLP:journals/tbbis/TerhorstFDKK21, DBLP:conf/icb/TerhorstFDKK20}.
Knowing the effect of individual attributes on the gender gap helps understand the causes of the accuracy disparity between genders. However, potential correlations and combinations of these attributes could strongly affect the interpretation.

\textit{In this work}, we fill this gap with three main contributions.
First, we propose a set of tools to decorrelate and aggregate facial attributes, enabling more objective reasoning about the gender gap.
Second, we adapt specialized metrics to enable quantification of the effect of facial attributes on gender fairness in relative and absolute terms.
Lastly, we present a novel unsupervised joint investigation framework to reliably identify attribute combinations that minimize the gender gap.
Our experiments begin with a set of 40 non-demographic attributes that describe elements of face image appearance, such as hair color, facial hair, or facial expression.
A systematic analysis was conducted to investigate the impact of demographic attributes on the difference in male and female face recognition accuracy. The results demonstrated that the gender disparity in face recognition accuracy is effectively eliminated when the test sets are balanced on a small number of relevant non-demographic attributes. This suggests that the observed discrepancy in face recognition accuracy between male and female cohorts is more accurately attributed to social norms surrounding male and female appearance, rather than to biological factors or inherent biases in deep neural networks. In other words, these appearance-driven differences create the \textit{illusion} of gender bias, even though the underlying models are not inherently biased towards specific gender.

In contrast to previous works, the proposed contributions mainly differ in three core aspects:

\begin{enumerate}
    \item \textbf{Large-scale Attribute Analysis} - We consider a wide variety of $40$ facial characteristics in this work, significantly more than most comparable studies.
    \item \textbf{Attribute Decorrelation} - Unlike previous works, we propose a facial attribute decorrelation toolchain, meant to increase the expressiveness of our results while mitigating the probability of correlations distorting the outcomes.
    \item \textbf{Unsupervised Joint Investigation Framework} - While previous works analyzed only individual attributes separately, we devise a completely novel approach to jointly investigate the impact of combinations of non-demographic attributes on gender bias. 
\end{enumerate}

\section{Related Work}
Gender bias in face recognition was first reported in the 2002 Face Recognition Vendor Test (FRVT)~\cite{phillipsFirstGenderDifferenceReport2003}.
Before deep learning algorithms became popular, \cite{phillipsPreDLGenderBias2005, beveridgePreDLGenderBias2009, luiPreDLGenderBias2009, grotherPreDLGenderBias2011, klarePreDLGenderBias2012} already conducted experiments with multiple face recognition algorithms and consistently reported lower accuracy for female cohorts than for male ones.
For deep-learning based face recognition, research increasingly focused on investigating and mitigating this bias.

\textit{\textbf{Investigating Gender Bias}}: In recent years, learning-based face recognition algorithms and datasets have led to remarkable performance, such as $>$99.8\% accuracy on the LFW benchmark~\cite{huangLFW2008}. 
However, this improvement in accuracy did not eliminate the accuracy disparity on gender~\cite{best-RowdenDLGenderBias2015, best-RowdenDLGenderBias2018, wuBalancedDataset2023}. 
A common assumption is that this gap is caused by female under-representation in training data, but \textcite{albieroHowDoesGender2020} reported that balancing male and female identities in the training data does not yield balanced accuracy on the test data.
\textcite{wuBalancedDataset2023} proposed a bias-aware dataset controlling head pose, image quality, and brightness of the images. Even after reducing these non-gender-related factors, they still observed males achieving 4.86\% higher true positive rate (TPR) than females.
Researchers have also investigated the effects of image attributes associated with gender by social custom.
For example, several works \cite{uedaMakeup2010, dantchevaMakeup2012, guoMakeup2014, albieroAnalysisGenderInequality2020} show that variations in makeup, which are more commonly associated with female than male cohorts, tend to reduce similarity scores for genuine (same-identity) image pairs, making it harder for the recognition algorithm.
The role of facial hair, commonly associated only with males, has also been investigated \cite{wuFacialHairAttributeInDemographic2023, wuFacialHairPositionEffect2024, ozturkBeardSizeEffect2023}.
Results show that matching beard regions increases similarity scores, while dissimilar beard regions achieve the opposite effect.  
This effect is amplified with growing size of facial hair, and when positioned in central regions of the face (e.g., moustaches).
Researchers ~\cite{albieroFaceRecognitionSexist2020a, wuBalancedDataset2023, albieroGenderedDifferences2022} also observed that male faces are generally larger, and that balancing test data on face morphology decreases gender bias.
\textcite{bhattaGenderDifferenceHair2023} examined scalp-based hairstyle difference between males and females. 
After balancing the test data, the gender difference for genuine pairs is strongly reduced, but not for impostor pairs. 
They found that these attributes strongly correlate, and that many non-demographic attributes have significant effect on recognition performance. 
Previous studies have made efforts to analyze sources of gender bias, primarily by concentrating on individual attributes.
This ignores potential correlations and combinations between multiple attributes affecting the final interpretation.

\textbf{\textit{Mitigating Gender Bias:}} 
The fact that gender bias was reported in a number of investigations motivated research into how to mitigate such bias.
A popular strategy in this context is to modify the training strategy or the network structure without changing the data.
\textcite{terhorstPostcomparisonMitigationDemographic2020} introduced an unsupervised fairness score normalization to reduce gender bias, which was refined by FALCON \cite{10943725} to improve score normalization through locally optimal feature adjustments and achieve stronger fairness-accuracy trade-offs.
\textcite{gongAdversarialDebias2020} presented a de-biasing adversarial network (DebFace), which learns how to extract disentangled, unbiased features for face recognition and demographic estimation, as a means to mitigate demographic accuracy disparities.
\textcite{ParkVAEDebias2021} introduced the Fairness-aware Disentangling Variational Auto-Encoder (FD-VAE), which disentangles the target attribute latent, protected attribute latent, and mutual attribute latent to mitigate the performance bias on gender and age.
\textcite{dharMitigateBias2021} proposed a Distill and De-bias (D\&D) structure to force a network to attend to similar face regions, irrespective of the attribute category. This approach is reported to reduce bias based on skintone.
\textcite{dharAdversarialDebias2021} proposed a descriptor-based adversarial de-biasing approach aimed at reducing gender and skintone information to minimize bias while maintaining high performance.
In conclusion, these works have the potential to narrow the gender gap. 
However, they approach the problem of mitigating unequal accuracy across gender without an understanding of the causes of unequal accuracy.
With a deeper understanding of the underlying causes, it may be possible to develop more effective solutions.

\textbf{\textit{This Work:}}
This paper is the first to approach the task of identifying facial characteristics accounting for gender bias in an entirely unsupervised manner.
While previous work manually identified probable causes before testing their effect on bias, this work takes a different perspective. By focusing on minimizing the gender gap as our primary optimization goal, we aim to identify those subsets of a large repository of decorrelated non-demographic attributes responsible for increased fairness as a by-product.
By testing face recognition models in this way, we believe that the identified subsets can be linked to the observed gender bias with a high degree of confidence yet minimal assumptions.

\section{Methodology}
This work explores how specific combinations of non-demographic facial attributes, shared across gender in the test data, minimize gender disparity.
Our methodology relies on three core concepts.
First, tailored fairness metrics are necessary to reliably assess the degree of gender bias in face recognition accuracy, presented in \cref{subsec:fairness-metric}.
Next, an efficient solution to forming attribute combinations and the corresponding test datasets is needed.
Our solution to this is detailed in \cref{subsec:forming-combinations}.
Last, we introduce a novel method of decorrelating facial attributes by clustering in \cref{subsec:attribute-decorrelation}, enabling more expressive and generalizable results.
Combining these techniques, we are able to project gender bias onto the presence or absence of a few facial features.

\subsection{Fairness Metric}
\label{subsec:fairness-metric}
Generally, fairness metrics express the equitability of a system w.r.t. specific groups.
For this work, we focus on equitable accuracy of face recognition for male and female cohorts.

\subsubsection{iGARBE}
\label{subsubsec:metric-igarbe}
We assume the notion of fairness in face biometrics as presented in \cite{pereiraFairnessBiometricsFigure2022}.
A fair FRS yields the same false non-match rate $\mathrm{FNMR}\mleft(\tau\mright)$ 
at a given false match rate $\mathrm{FMR}_{x}\mleft(\tau\mright)$ for the considered subject groups.
Here, $\tau$ corresponds to the decision threshold where $\mathrm{FMR}$ is equal to a chosen operational point $x$ (e.g., $10^{-3}$ as proposed in \cite{frontexBestPracticeTechnical2015}).
Multiple fairness metrics exist that take up on this concept \cite{pereiraFairnessBiometricsFigure2022, grotherDemographicDifferentialsFace2021}, a recently presented one being GARBE \cite{howardEvaluatingProposedFairness2022}.
It addresses accuracy and interpretability issues of previously proposed metrics, and is thus the most reliable and precise option known to us.

GARBE is based on an adjusted version of the well-known Gini coefficient $G$ \cite{deltasSmallSampleBiasGini2003a}.
GARBE and the adjusted version of $G$ are defined as:
\begin{align}
	&\begin{aligned}
		&\mathrm{GARBE}\mleft(\tau\mright) = \alpha A\mleft(\tau\mright) + \left(1-\alpha\right)B\mleft(\tau\mright)\quad\,\,\, &&\text{where}\\
		&A\mleft(\tau\mright) = G_{\mathrm{FMR}_{\tau}} &&\text{and}\\
		&B\mleft(\tau\mright) = G_{\mathrm{FNMR}_{\tau}}
	\end{aligned} \label{eq:garbe-total} \\
	&G_x = \left(\frac{n}{n-1}\right) \left(\frac{\sum _{i=1}^{n} \sum_{j=1}^{n} \lvert x_i - x_j \rvert}{2n^{2}\bar{x}}\right)\,\,\,\,\,\forall d_i, d_j \in D \label{eq:garbe-base}
\end{align}
respectively.

We adapt the original GARBE to reflect the intuition that a perfectly fair system have a score of \num{1} and a perfectly unfair system a score of \num{0}.
The original GARBE behaves inversely to this.
Since it is bounded to $\lbrack 0,1\rbrack \subset \mathbb{R}$, we compute the inverse GARBE (iGARBE) as
\begin{equation}
    \label{eq:igarbe}
	\mathrm{iGARBE}\mleft(\tau\mright) = 1-\mathrm{GARBE}\mleft(\tau\mright)
\end{equation}
Accordingly, iGARBE will be used as the fairness metric of choice in all following sections of this work.

\subsubsection{Contextualized Fairness (CoFair)}
\label{subsubsec:metric-contextualization}
The iGARBE metric, as well as the Gini coefficient it is based on, are reliable absolute measures of fairness.
However, their results provide no information about how much better the fairness scores computed on one set of attributes are relative to the scores computed on another set of attributes.
We therefore propose the computation of contextualized fairness, or CoFair in short, as an additional figure of scrutiny.
Given an FRS-specific, estimated probability density of iGARBE scores $f$ and an iGARBE score $s$, $\mathrm{CoFair}\mleft( s \mright)=p$ simply reflects the cumulative probability $p$ of any other iGARBE score $S$ being smaller than $s$.
This definition is presented in \cref{eq:igarbe-cdf} in more rigor.
\begin{equation}
	\label{eq:igarbe-cdf}
	\begin{aligned}
		&\mathrm{CoFair}\mleft( s\mright) = F_{S}\mleft( s \mright) = P\mleft(S \leq s \mright) = \int_{-\infty}^{s} f_{S}\mleft( t \mright)\,dt
	\end{aligned}
\end{equation}
Put simply, CoFair expresses the expected fraction of iGARBE scores, computed on varying sets of attributes, that are smaller than or equal to the given score (and thus, the given set of attributes) for a fixed FRS.
Therefore, the higher the CoFair value of such a score, the better.

\subsection{Forming Fair Attribute Combinations}
\label{subsec:forming-combinations}
We define an attribute combination as a nonempty set of labeled non-demographic attributes.
Throughout this work, the terms \emph{labeled (non-demographic) attributes} and \emph{assignments} are used interchangeably.
We use the MAAD-Face database \cite{terhorstMAADFaceMassivelyAnnotated2021} as a reference for available non-demographic attributes.
It comprises 3.3M images of over 9k distinct individuals, with each image being annotated with  \num{7} demographic and \num{40} non-demographic attributes (related to, e.g., hair or accessories).
Accordingly, the accommodating ternary labels are taken from MAAD-Face as well.
However, for the combinations, we limit the space of available labels to only positive (\num{1}) and negative (\num{-1}), omitting unclear (\num{0}) labels to increase expressiveness.
Positive labels indicate that the given attribute must be present in the corresponding faces.
Conversely, negative labels indicate the specific absence of a given attribute.

Forming fair attribute combinations requires measuring combinations' iGARBE scores.
To this end, we use a custom database of comparisons of annotated image templates.
The annotated templates are provided by MAAD-Face's annotated version of VGGFace2 \cite{caoVGGFace2DatasetRecognising2018a}, relevant details are provided in \cref{subsec:databases}.
The comparison database is then filtered for pairs of same-gender templates whose annotations conform to the requirements set by the assignment combination in question, as described in section \cref{subsubsec:filtering}.
In consideration of the focus of this work, the resulting subset is then further decomposed into male and female subgroups, respectively.
To eliminate the chance of sample-size disparities negatively impacting further analyses, the subgroups' sets are further sampled (see \cref{subsubsec:sampling} for more details).
Throughout this paper, we refer to the process of filtering and sampling given an assignment combination as ``equalizing'' or ``equalization'', describing the forced sharing of an attribute subset across genders.
We compute the iGARBE scores over those samples to measure the effect of the attribute combinations in question, linking the results directly to the combination.
Accordingly, we call an attribute combination \emph{fair} if its iGARBE score approaches \num{1} sufficiently closely at a fixed FMR.
Similarly, we call it \emph{fairness-increasing} if its iGARBE score increases relative to a provided baseline at a fixed FMR.

\begin{figure}[t]
	\centering
	\includegraphics[width=\linewidth]{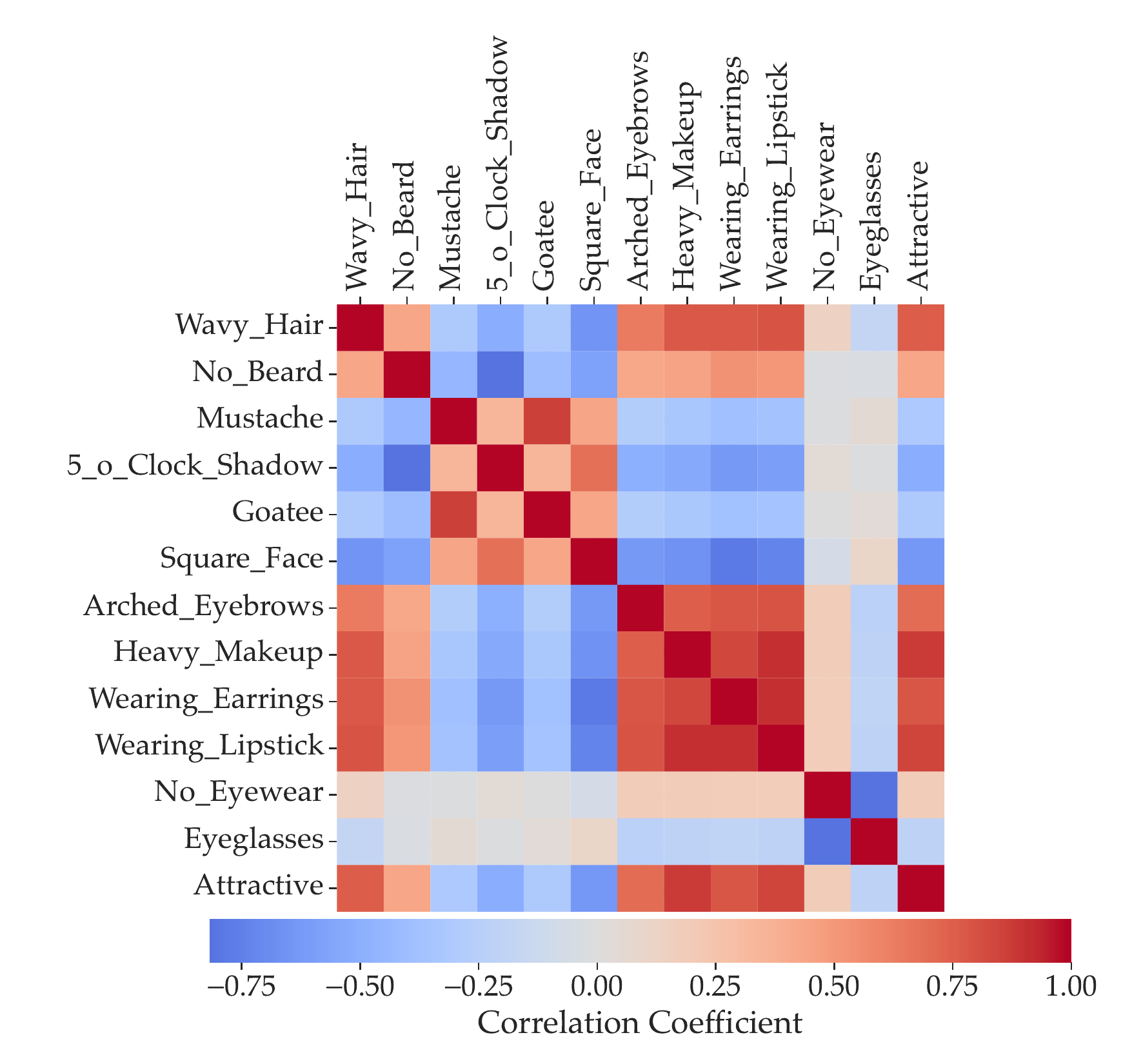}
	\caption{\textbf{Attribute annotation correlations} - The correlations are computed using the Pearson coefficient. The depicted attributes are selected such that the \num{15} highest absolute, i.e., positive or negative, correlations are visible. As can be seen, very strong correlations exist between various attributes, indicating that they are not statistically independent.}
	\label{fig:correlation-analysis-top-15}
\end{figure}

Given these concepts, the question remains how combinations that are at least fairness-increasing and ideally fair can be determined.
It might seem trivial to combine those labeled attributes, which are each fairness-increasing or even fair by themselves.
However, this assumes that the provided attributes are statistically independent.
To investigate this assumption, we compute the Pearson correlation between all pairs of attributes.
A relevant excerpt of the results is shown in \cref{fig:correlation-analysis-top-15}.
It depicts all correlations between those attributes that are featured in the \num{15} highest \emph{absolute} pairwise correlations.
As can be seen, correlations with absolute values of \num{0.75} and higher exist.
Clearly, the attributes at hand are not statistically independent.
Therefore, we have to form and examine combinations as standalones, discarding the individual effects of attributes they comprise.

Under this assumption, however, the search space becomes very large.
As previously elaborated, MAAD-Face assigns each image template with one of three values for each of \num{40} non-demographic attributes.
While we limit ourselves to only use explicitly positive or negative labels for attributes that we combine, the remaining annotations are left undefined.
Thus, we can approximate the magnitude of possible assignment combinations to be around $3^{40}-1 \approx 1.2 \cdot 10^{19}$.
Computing the effects of all these combinations on fairness is practically impossible in limited time.

Hence, a guided approach that allows to effectively yet efficiently navigate the search space is needed.
We therefore propose a greedy attribute-combining algorithm similar to breadth-first search.
At its heart, it adds the $n \in \mathbb{N}$ most relevant, fairness-increasing assignments to a given node in the search tree, starting with the root node.
The relevance of assignments is determined via a custom ranking metric elaborated on in \cref{subsubsec:ranking}.
This process is repeated for each node in the tree at a given depth $d \in \mathbb{N}$ until a pre-defined depth limit $d_{max} \in \mathbb{N}$ is reached.
Upon completion, each branch in the tree represents a fairness-increasing assignment combination.
The corresponding algorithm is presented in \cref{alg:assignment-combination-forming}.
\begin{algorithm}[t]
	\caption{- Assignment Combination Forming}
	\label{alg:assignment-combination-forming}
	\begin{algorithmic}[1]
		\Require $\mathcal{T}$, $\mathcal{A}^{l}$, $f_{0}$, $d_{max}$, $n$
		\Ensure $\mathcal{T}'$
		\State $\mathcal{T}' \gets \mathcal{T}$
		\ForAll{$d \in \lbrack 0, d_{max} \rbrack \subset \mathbb{N}$}
			\State $\mathcal{N} \gets$ \Call{NodesAtDepth}{$d$, $\mathcal{T}'$}
			\ForAll{$\nu \in \mathcal{N}$}
				\State $\mathcal{M}_{\nu} \gets \lbrace \rbrace$
				\State $\mathcal{B} \gets$ \Call{BranchToNode}{$\nu$, $\mathcal{T}'$}
				\ForAll{$A^{l} \in \mathcal{A}^{l}$}
					\State $\mathcal{C} \gets \left\lbrace A^{l} \right\rbrace \cup \mathcal{B}$
					\State $m_{A^{l}} \gets$ \Call{MetricsOfCombination}{$\mathcal{C}$}
					\If{\Call{Fairness}{$m_{A^{l}}$} $> f_{0}$}
						\State $\mathcal{M}_{\nu} \gets \mathcal{M}_{\nu} \cup \left\lbrace \left(m_{A^{l}}, A^{l} \right) \right\rbrace$
					\EndIf
				\EndFor
				\State $\mathcal{A}_{r}^{l} \gets$ \Call{RelevantAssignments}{$n$, $\mathcal{M}_{\nu}$}
				\State \Call{AddToNode}{$\mathcal{A}_{r}^{l}$, $\nu$, $\mathcal{T}'$}
			\EndFor
		\EndFor
		\State \Return $\mathcal{T}'$
	\end{algorithmic}
\end{algorithm}
In the following, the previously referenced components supporting this approach, namely filtering, sampling, and ranking, are explained in detail.

\subsubsection{Filtering}
\label{subsubsec:filtering}
In the proposed approach, a crucial step is filtering all available template comparisons for those in which both partners fulfill the requirements regarding the presence or absence of attributes.
These requirements, as previously elaborated, are set by the assignment combination in question.

Given our experimental workflow and tools described in \cref{sec:experimental-setup}, we approach this problem in three steps.
First, filter the labeled individual templates for those of interest, i.e., those annotated with the assignments of the currently inspected combination.
Second, semi-join \cite{bernsteinUsingSemiJoinsSolve1981} the indices of these filtered templates, in this case the primary keys, on the comparisons dataset.
Since each comparison is comprised of two paired templates, this join is executed on the index of each partner, respectively.
Last, intersect the results to retrieve only those comparisons in which \emph{both} partners match the given specifications.

\subsubsection{Sampling}
\label{subsubsec:sampling}
We sample the filtered comparisons to mainly rule out sample size disparities as root causes for bias.
In particular, the sampling process enforces equal sample counts across male and female subgroups for each evaluated attribute combination. 
This ensures that the resulting comparisons are balanced with respect to gender, even though the original dataset is not.
Simultaneously, sampling also prevents small-sized outliers to distort any results.
It is therefore crucial to take these two aspects into account to ultimately enhance informational value.

To this end, we present a custom sampling method.
We assemble $\gamma$ sets of comparison samples per subgroup, such that all sets contain the same number of samples.
These samples are chosen uniformly at random from the available comparisons, i.e., those resulting from filtering.
Additionally, each such set must adhere to a sample ratio $\rho_{s}=\tfrac{1}{\beta}$, with $\beta$ dictating how many imposter comparisons per genuine comparison the set must comprise.
$\lambda_{g}$ and $\lambda_{i}$ are used as lower boundaries, describing the minimum number of genuine and imposter samples that must be contained in each such set.
Note that we ensure that for each subgroup, $\bigcap^{\gamma} S_{g}^{\left(\cap\emptyset\right)} = \emptyset$, i.e., that the available genuine samples are uniquely distributed across all $\gamma$ sets.
However, we disregard this requirement for imposter samples.
Since we compute the FNMR at a fixed FMR for our results, it is critical to minimize the risk of duplicated outlier genuine comparisons affecting the FNMR.
As there are more imposter comparisons at our disposal in the first place, which by their nature are also more noisy, it is not necessary to enforce these rules w.r.t. the FMR.

Ultimately, \cref{alg:sample} therefore yields $\gamma$ fixed-sized, bounded sample sets of genuine and imposter comparisons contained in $\mathcal{G}_{g}^s$, $\mathcal{G}_{i}^s$ for all given genuine and imposter comparisons per subgroup provided through $\mathcal{G}_{g}$, $\mathcal{G}_{i}$.
\begin{algorithm}[t]
	\caption{- Comparison Sampling}
	\label{alg:sample}
	\begin{algorithmic}[1]
		\Require $\mathcal{G}_{g}$, $\mathcal{G}_{i}$, $\rho_{s}$, $\lambda_{g}$, $\gamma$
		\Ensure $\mathcal{G}_{g}^s$, $\mathcal{G}_{i}^s$
		\LineComment{Determine max count of samples equal across genders}
		\State $c_{g}^{\left( + \right)} \gets$ \Call{MaxEqualSampleCount}{$\mathcal{G}_{g}$}
		\State $c_{i}^{\left( + \right)} \gets$ \Call{MaxEqualSampleCount}{$\mathcal{G}_{i}$}
		\State $c_{g}^{(\cap\emptyset)} \gets \left\lfloor\frac{c_{g}^{\left( + \right)}}{\gamma}\right\rfloor$
		\State $\lambda_{i} \gets \left\lfloor\frac{\lambda_{g}}{\rho_{s}}\right\rfloor$
		\If{$\left(c_{g}^{(\cap\emptyset)} < \lambda_{g}\right)\vee\left(c_{i}^{\left( + \right)} < \lambda_{i}\right)$}
			\State \Return $\emptyset$, $\emptyset$
		\EndIf
		\State $c_{g}$, $c_{i} \gets 0$
		\LineComment{Adjust sample sizes to respect sample ratio}
		\If{$\left(c_{g}^{(\cap\emptyset)}\cdot\rho_{s}\right) > c_{i}^{\left( + \right)}$}
			\LineComment[1]{Keep all imposter samples}
			\State $c_{g} \gets \left\lfloor c_{i}^{\left( + \right)}\cdot\rho_{s}\right\rfloor$
			\State $c_{i} \gets c_{i}^{\left( + \right)}$
		\Else
			\LineComment[1]{Keep all genuine samples}
			\State $c_{g} \gets c_{g}^{(\cap\emptyset)}$
			\State $c_{i} \gets \left\lfloor\frac{c_{g}^{(\cap\emptyset)}}{\rho_{s}}\right\rfloor$
		\EndIf
		\State $\mathcal{G}_{g}^s$, $\mathcal{G}_{i}^s \gets \lbrace\rbrace$, $\lbrace\rbrace$
		\ForAll{$\mathcal{S}_{g} \in \mathcal{G}_{g}$, $\mathcal{S}_{i} \in \mathcal{G}_{i}$}
			\State $\mathcal{S}_{g}^{(\cap\emptyset)} \gets$ \Call{NDisjointSampleGroups}{$\gamma$, $c_{g}$, $\mathcal{S}_{g}$}
			\State $\mathcal{S}_{i}^s \gets$ \Call{NSampleGroups}{$\gamma, c_{i}$, $\mathcal{S}_{i}$}
			\State $\mathcal{G}_{g}^s \gets \mathcal{G}_{g}^s \cup \left\lbrace\mathcal{S}_{g}^{(\cap\emptyset)}\right\rbrace$
			\State $\mathcal{G}_{i}^s \gets \mathcal{G}_{i}^s \cup \left\lbrace\mathcal{S}_{i}^s\right\rbrace$
		\EndFor
		\State \Return $\mathcal{G}_{g}^s$, $\mathcal{G}_{i}^s$
	\end{algorithmic}
\end{algorithm}

\begin{figure}[t]
	\centering
	\includegraphics[width=\linewidth]{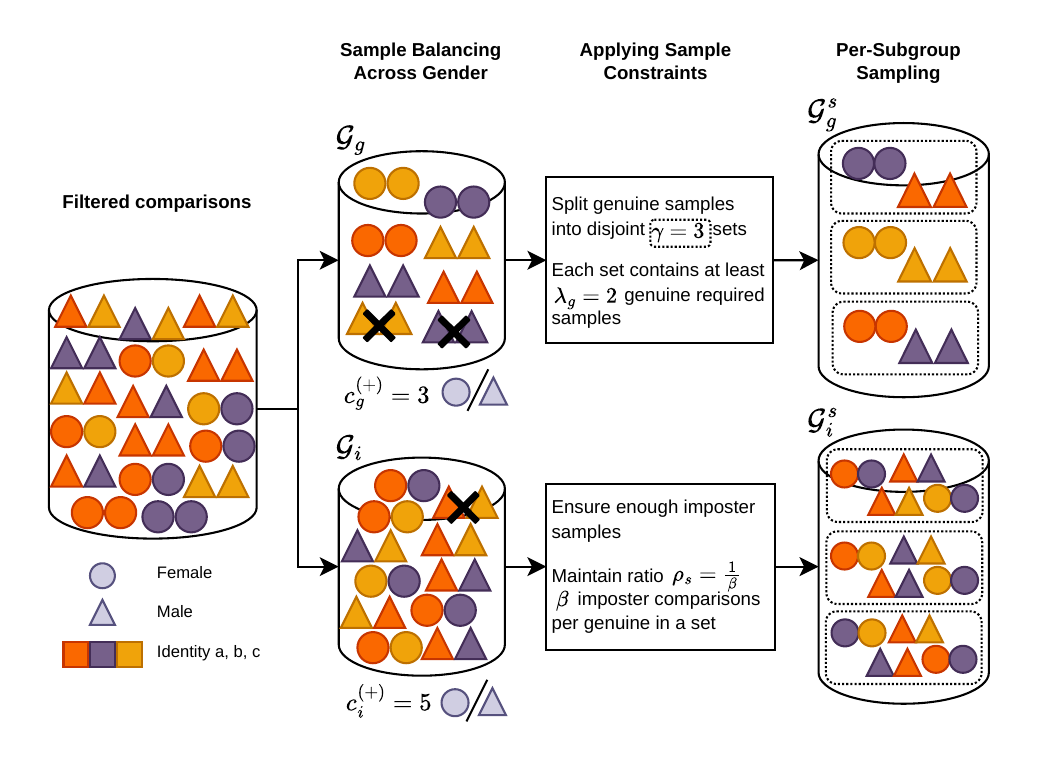}
	\caption{\textbf{Comparison Sampling Visualization} - Filtered comparisons are separated into genuine and imposter groups, which are further balanced across gender subgroups. Constraints are then applied while partitioning the comparisons into $\gamma$ fixed-size sets, ensuring disjoint genuine samples and a defined ratio of reusable imposters, forming $\mathcal{G}_g^s$ and $\mathcal{G}_i^s$ for robust evaluation.}
	\label{fig:algorithm2-visualization}
\end{figure}

In this work, we set $\rho_{s} = \tfrac{1}{5}$ to balance computational complexity and the number of samples taken into consideration.
For the same reasons as well as the advantage of robust computations of means and standard deviations, we choose $\gamma = 3$.
Lastly, $\lambda_{g}=\tfrac{1}{\mathrm{FMR}}$ is set to consistently ensure an adequate number of samples given the level of scrutiny the system is put under due to the FMR and the corresponding decision thresholds. 
An overview of the sampling process is shown in Figure \ref{fig:algorithm2-visualization}. 
The method constructs $\gamma$ balanced comparison sets per subgroup by enforcing minimum sample constraints and a fixed genuine-to-imposter ratio,  with genuine comparisons assigned disjointly and imposter comparisons sampled with replacement.

\subsubsection{Ranking}
\label{subsubsec:ranking}
The proposed approach progressively forms fairness-increasing assignment combinations.
As previously described, this involves extending the corresponding search tree with the $n \in \mathbb{N}$ most relevant assignments for each node at each level.
Determining these assignments requires two steps: first ranking them, then pruning them for the results to reflect the desired top-n characteristics.
While pruning is a straightforward process, ranking is more intricate.
We will therefore present the corresponding metric devised for this work in the following.

In designing the ranking metric, the goal is to characterize the effects of assignments w.r.t. three core factors.
First, the genuine sample retention is considered.
Compared to the originally available number of samples, the currently inspected assignment combination should retain as many as possible.
Second, we inspect the assignment combination's effect on the verification error.
It should not increase the overall system's verification error, while decreases should be rewarded.
Consistency should neither be penalized nor rewarded.
Third, in line with this work's objective, the assignment combination in question should increase fairness.
The greater the increase, the higher the reward.
Decreases are penalized.
Consistency is not minded.
All of these factors are to be \emph{weighted equally} in the evaluation.

For the implementation of these requirements, we use the well-known sigmoid function $\sigma$ as a baseline, primarily to leverage its $\lbrack 0, 1 \rbrack$-bounded properties.
Consequently, we define our ranking metric as shown in \cref{eq:relevance}.
\begin{align}
	&R = \tfrac{1}{3}\left(R_{s} + R_{p} + R_{f}\right) \,\, \text{with} \label{eq:relevance} \\
	&R_{s}\mleft(n_{i}, \mathrm{FMR}\mright) = \sigma\mleft(\left(-\mathrm{FMR} \cdot n_{i} \cdot \left(\ln\mleft(4\mright) - \mu\right)\right)-\mu\mright) \label{eq:sample-retention} \\
	&R_{p}\mleft(\mathrm{FNMR}_{i}, \mathrm{FNMR}_{0}\mright) = \sigma\mleft(\lambda\left(\mathrm{FNMR}_{0} - \mathrm{FNMR}_{i}\right)\mright) \label{eq:total-fnmr} \\
	&R_{f}\mleft(f_{i}, f_{0}\mright) = \sigma\mleft(\omega\left(f_{i} - f_{0}\right)\mright) \label{eq:fairness}
\end{align}
Equations \ref{eq:sample-retention}, \ref{eq:total-fnmr}, and \ref{eq:fairness} correspond to the previously described core factors: $R_s$ to genuine sample retention, $R_p$ to verification error, and $R_f$ to fairness.
They are structured such that the following holds.
\begin{alignat*}{2}
	&R_{s}\mleft(\tfrac{1}{\mathrm{FMR}},\mathrm{FMR}\mright) \,\, &&=0.2 \\
	&R_{p}\mleft(\mathrm{FNMR}_0,\mathrm{FNMR}_0\mright) \,\, &&=0.5 \\
	&R_{f}\mleft(f_0,f_0\mright) \,\, &&=0.5
\end{alignat*}
In all formulas, variables indexed with \num{0} represent baseline values, while those indexed with $i$ represent values resulting from evaluating the currently inspected assignment combination.
The number of retained genuine samples is represented through $n$, the current iGARBE score through $f$.
The hyperparameters $\mu,\lambda,\omega \in \mathbb{R}_+$ determine the slope, i.e., how quickly higher scores should be awarded.
We desire a moderate scoring behaviour to neither reward minor improvements too highly nor penalize lightly deteriorated values too drastically.
Moreover, $R_s$, $R_p$, and $R_f$ should behave similarly over relative, semantically equivalent ranges of their relevant domain.
To estimate generic hyperparameters for this work, we assume there to be at max $10^7$ genuine samples, $\mathrm{FNMR}_0=0.1$, and $f_0=0.9$.
A suitable estimation under these constraints is $\mu=1.3865$ and $\lambda=\omega=4$. 

\label{subsubsec:incremental-clustering}
\begin{algorithm}[t]
	\caption{- Correlated Attribute Clustering}
	\label{alg:cluster}
	\begin{algorithmic}[1]
		\Require $\mathcal{A}, i_{max}$
		\Ensure $\mathcal{A}_{\indep}$
		\State $\mathcal{A}_{\indep} \gets \mathcal{A}$
		\ForAll{$\_ \in \lbrack 0, i_{max}-1 \rbrack \subset \mathbb{N}$}
			\State $\mathcal{R} \gets \emptyset$
			\ForAll{$A_{\indep} \in \mathcal{A}_{\indep}$}
				\State $\Delta\mathcal{A}_{\indep} \gets \mathcal{A}_{\indep} - \lbrace A_{\indep} \rbrace$
				\ForAll{$\Delta A_{\indep} \in \Delta\mathcal{A}_{\indep}$}
					\State $r \gets \lvert$ \Call{Pearson}{$A_{\indep}, \Delta A_{\indep}$} $\rvert$
					\State $\mathcal{R} \gets \mathcal{R} \cup \left\lbrace \left( r, A_{\indep}, \Delta A_{\indep} \right) \right\rbrace$
				\EndFor
			\EndFor
			\State $\_, A_{\indep,1}, A_{\indep,2} \gets \max_{r} \mathcal{R}$
			\State $\mathcal{A}_{\indep} \gets \left ( \left ( \mathcal{A}_{\indep} - A_{\indep,1} \right ) - A_{\indep,2} \right ) \cup \lbrace A_{\indep,1}, A_{\indep,2} \rbrace$
		\EndFor
		\State \Return $\mathcal{A}_{\indep}$
	\end{algorithmic}
\end{algorithm}

\subsection{Decorrelating Attribute Annotations}
\label{subsec:attribute-decorrelation}
As shown in \cref{fig:correlation-analysis-top-15}, the attributes used for annotations in MAAD-Face are highly correlated w.r.t. Pearson correlation.
For this work, this is problematic as it restricts the exploratory capabilities of the approach presented in \cref{subsec:forming-combinations}.
This is because of the limited branch depth.
Assume the discovery of a fairness-increasing assignment combination made up of only correlating attributes.
Then, this combination of attributes relating to similar underlying characteristics occupies one branch of the tree.
Uncovering such correlated fairness-increasing assignment combinations undoubtedly is of high interest.
However, the focus of this work is to provide as broad of an overview over likely causes of gender bias as possible.
Not limiting the effects of these correlations would interfere with this objective by reducing the expressiveness and informational value of the presented method and its results.
Therefore, we devise a decorrelation method to create clusters of correlated attributes such that the underlying attributes can be treated as one.
It involves two steps.
First, clusters themselves must be formed.
This process is described in \cref{subsubsec:incremental-clustering}.
Second, the clusters need to be harmonized to reduce the overall correlation.
Details are provided in \cref{subsubsec:cluster-harmonization}.

\subsubsection{Incremental Clustering}
We intend clusters to comprise attributes that significantly correlate with each other w.r.t. Pearson correlation.
This implies both positive and negative correlations, as both cases need to be accounted for in further steps.
Thus, we incrementally create clusters based on the pairwise absolute Pearson correlation.
In the following, clusters may contain an arbitrary number of attributes in the range $\lbrack 1, \vert \mathcal{A} \rvert \rbrack \subset \mathbb{N}$, i.e., individual attributes are also viewed as clusters.
The corresponding algorithm is presented in \cref{alg:cluster}.

\begin{figure}[t]
	\centering
	\includegraphics[width=\linewidth]{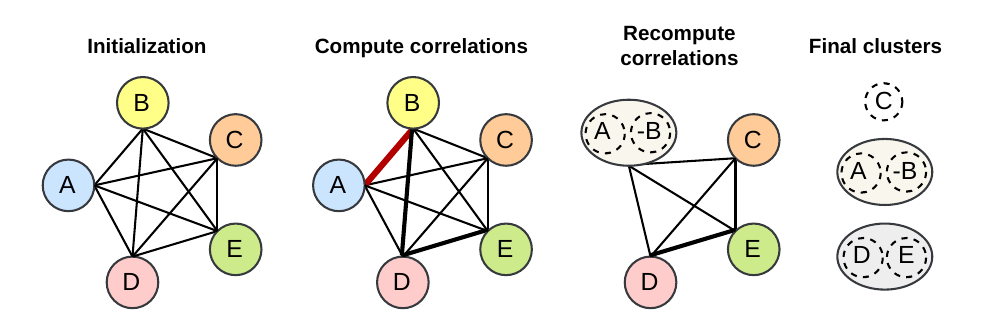}
	\caption{\textbf{Correlated Attribute Clustering Visualization} - This process incrementally merges the most strongly correlated attribute pairs based on pairwise absolute Pearson correlation. At each iteration, the pair with the highest absolute correlation (including negative relationships represented via sign inversion, e.g., $-B$) is selected, the attribute set is updated, and correlations are recomputed, progressively forming clusters of related attributes.}
	\label{fig:algorithm3-visualization}
\end{figure}
Given the initial set of attribute annotations $\mathcal{A}$ as provided by MAAD-Face, it merges the two most correlated clusters in each of the $i_{max} \in \mathbb{N}$ iterations, with $0 \leq i_{max} \leq \lvert \mathcal{A} \rvert$.
To this end, the pairwise absolute Pearson correlation of all clusters, $\mathcal {R}$, is determined.
If two clusters contain more than one attribute, we calculate the mean pairwise absolute Pearson correlation between all pairs of attributes that are not part of the same cluster.
Subsequently, we merge the pair of clusters $\lbrace A_{\indep, 1}, A_{\indep, 2} \rbrace$ with the maximum out of all gathered absolute Pearson correlations.

As determining an optimal $i_{max}$ is not generalizable, the required process and corresponding results are detailed in \cref{subsec:correlation-mitigation}. 
A visualization of the incremental clustering algorithm is shown in Figure \ref{fig:algorithm3-visualization}.
Starting from singleton attributes, at each iteration, the pair with the highest absolute Pearson correlation is identified and merged into a cluster.
By considering absolute correlation values, both positive and negative relationships contribute to cluster formation, while clusters progressively grow through iterative updates of the attribute set.

\subsubsection{Cluster Harmonization}
\label{subsubsec:cluster-harmonization}
As stated previously, we intend to use the clusters resulting from the iterative decorrelation methodology as if they were standalone attributes.
However, doing so is not straightforward.
Since we use the absolute correlation as a baseline metric, we consciously encourage contradicting but correlating attributes to be clustered in particular.
This imposes a problem when trying to assign labels to a cluster as is necessary for the presented approach for forming assignment combinations.
Assuming inverse underlying semantics of two clustered attributes $A_1$ and $A_2$, assigning $\lbrace A_1, A_2 \rbrace = 1$ would simultaneously require the presence and absence of the same facial characteristic.
Subsequent filtering for samples that accord with this predicate would yield no results, rendering our clusters unusable.
Consequently, we harmonize the clusters.

In the case of this work, harmonizing entails ensuring that clusters only contain positive correlations without adding or removing any attributes.
Thereby, the most transparent behaviour is ensured.
Assigning positive labels to a cluster reflects the specific presence of the underlying attributes, while assigning negative labels does the inverse.
To facilitate these characteristics, first, the two most negatively correlating attributes of a given cluster are determined.
If there are no negative correlations, we do not need to harmonize the cluster.
Otherwise, one of these two attributes is used as reference.
All other attributes in the cluster will then be inverted, if necessary, such that their correlation with the reference attribute will be exclusively positive.
Note that this results in exclusively positive \emph{pairwise} correlations between \emph{all attributes} in the given cluster.

To correctly integrate this behaviour with the rest of our work, the inversion of an attribute leads to three actions being taken.
First, adding the prefix ``Not'' to intuitively reflect the semantics of the change.
Second, all samples with a non-zero annotation of the now inverted attribute have to be inverted as well to guarantee uniformity across the given datasets.
Third, the correlations must be updated accordingly.
By conforming to these steps, clusters can now be assigned labels as if they were mere attributes without any issues.

\section{Experimental Setup}
\label{sec:experimental-setup}
\subsection{Databases}
\label{subsec:databases}
The goal of this work is to provide an approach to explain gender bias in face recognition via combinations of non-demographic attributes.
To facilitate this task, the use of a large-size database with many high-quality attribute annotations is required.
It ensures unconstrained conditions and thus allows making expressive and generalizable statements.
Based on these specifications, we choose the publicly available MAAD-Face annotation database \cite{terhorstMAADFaceMassivelyAnnotated2021}.
It annotates each of the \num{3.3}M images of over \num{9}k distinct individuals of VGGFace2 \cite{caoVGGFace2DatasetRecognising2018a} with \num{47} ternary attributes.
Of those, \num{7} relate to demographic factors, namely gender, ethnicity, and age.
The remaining \num{40} relate to non-demographic facial features such as hair or accessories.
As the focus of this work is gender bias, we desire an acceptably balanced dataset w.r.t. gender to equally represent male and female individuals.
In VGGFace2 and, thus, MAAD-Face, about \qty{60}{\percent} of the images are of male individuals and the remaining approx. \qty{40}{\percent} of female subjects.
Moreover, the images feature a variety of head poses, thereby providing an even broader representation of recognition scenarios.
Lastly, MAAD-Face's attribute annotations are proven to have higher quality than comparable face annotation databases \cite{terhorstMAADFaceMassivelyAnnotated2021}.

\subsection{Face Recognition Models}
\label{subsec:fr-models}
For all experiments conducted in this work, we use templates created by two of the most popular face recognition models trained on two widely-used loss functions: ArcFace \cite{dengArcFaceAdditiveAngular2019} and FaceNet \cite{schroffFaceNetUnifiedEmbedding2015}.
Both models were prepared as described in \cite{terhorstComprehensiveStudyFace2022}.
In short, the pre-trained models for ArcFace\footnote{\url{https://github.com/deepinsight/insightface}} and FaceNet\footnote{\url{https://github.com/davidsandberg/facenet}} are based on ResNet-100 backbones and are trained on the MS1M database \cite{guoMSCeleb1MDatasetBenchmark2016}.
They are then applied to images that, for ArcFace, were pre-processed as described in \cite{guoStackedDenseUNets2018} and, for FaceNet, as described in \cite{kazemiOneMillisecondFace2014}.
The resulting templates are then compared using cosine similarity to enable identity verification. 
In addition to these two loss functions, a range of more recent face recognition approaches based on margin-based softmax losses have been proposed, such as AdaFace \cite{DBLP:conf/cvpr/Kim0L22} and CurricularFace \cite{DBLP:conf/cvpr/HuangWT0SLLH20}.
These methods follow the same underlying formulation as ArcFace, differing primarily in how the margin is adapted or weighted during training, while typically providing incremental improvements in performance. As such, they represent variations within the same family of margin-based approaches rather than fundamentally different modeling paradigms. Given that these methods share the same underlying formulation as ArcFace, similar trends in fairness behavior can be expected across these variants.

In light of this, we focus on models that cover substantially different design principles, rather than evaluating multiple closely related variants of the same loss formulation. 
The two selected face recognition models in this work differ in their training objectives and representation characteristics.
On one hand, FaceNet represents an earlier metric-learning approach, whereas ArcFace reflects a more recent margin-based formulation. 
Despite these differences, both models exhibit highly consistent patterns in fairness across attribute combinations, as shown in the results section. 
This suggests that the observed findings are not specific to a single architecture or loss function, but rather reflect more general properties of face representations.

\subsection{Metrics}
\label{subsec:metrics}
We report the results of our investigations in terms of FNMRs at decision thresholds corresponding to fixed FMRs.
These error rates are the international standard for biometric verification evaluation \cite{InformationTechnologyBiometric2021}.
Throughout our analyses, we set $\mathrm{FMR}=10^{-3}$, conforming to the recommendations of the European Border And Coast Guard Agency Frontex \cite{frontexBestPracticeTechnical2015}.
We measure the resulting FNMRs such that both total and differential verification performance can be inspected.
Thereby, effects on overall system performance as well as effects on male and female subgroups are quantified.

To assess fairness, we compute the iGARBE scores based on those metrics, as described in \cref{subsubsec:metric-igarbe}.
Doing so allows us to express the differential outcome, as per the definition of \textcite{howardEffectBroadSpecific2019}, across an arbitrary number of compared demographic groups in a single $\lbrack 0, 1 \rbrack$-bounded score.
Additionally, it enables the simple comparison of results in the context of a single face recognition model.
To also facilitate inter-model discussions about fairness, we contextualize these values with their underlying distribution characteristics through computing $\mathrm{CoFair}$ as detailed in \cref{subsubsec:metric-contextualization}.
Using the conjunction of these metrics provides us with the necessary tooling to approach the desired fairness analyses holistically.

Lastly, we also provide the number of genuine samples that meet the criteria dictated by the respective attribute filter predicate.
With the BioQuake uncertainty measure \cite{Fallahi2025OnTR}, we ensure that enough comparisons are provided to ensure significant outcomes. More importantly, we follow the 1\% rule to ensure that with 95\% confidence, the true error differs by no more than 1\% of the observed error. 
Although the high baseline quality of all results is programmatically enforced via strict sampling constraints as per \cref{subsubsec:sampling}, incorporating the size of the underlying genuine dataset aids in better understanding the impact of the findings w.r.t. their magnitude.

\subsection{Investigations}
\label{subsec:investigations}
We apply and evaluate the proposed methodology in multiple steps to investigate the combined effect of non-demographic attributes on fairness.
First, we assess the impact of progressive decorrelation to determine the most suitable setting for all further analyses.
Using the resulting attribute clusters, we then evaluate how ArcFace and FaceNet perform on samples reflecting the presence or absence of one or multiple of those clusters and their underlying attributes.
Specifically, we inspect the differences in error rates between male and female-labelled samples to assess the clusters' influence on gender fairness.
In this respect, we proceed in two stages.
First, we investigate the effect of individual clusters adhering to the filtering constraints we imposed.
Then, we subsequently re-perform our investigations once our proposed attribute combination approach was applied to those clusters.
In doing so, we emphasize the effectiveness of the second approach, which is the focus of this work.

\section{Results}
\subsection{Mitigating the Effects of Attribute Correlations}
\label{subsec:correlation-mitigation}
As seen in \cref{fig:correlation-analysis-top-15}, various attributes provided by MAAD-Face \cite{terhorstMAADFaceMassivelyAnnotated2021} are highly correlated with each other.
If these correlations are not taken into account or mitigated, misinterpretations or limited expressiveness of our approach's results may be the consequence.
For instance, a study on DeepFake detection \cite{11099163} demonstrated that these attribute correlations could lead to misleadingly high performances on widely-used benchmarks, underscoring their significance and interpretation.
To prevent such issues, we apply the iterative clustering-based decorrelation approach presented in \cref{subsec:attribute-decorrelation} to group together highly correlated attributes and thus, to form more uncorrelated attributes.
To determine the optimal clustering, we define three core criteria in the following.

First, since the overall objective is to retrieve decorrelated clusters, it is desirable to achieve a \textbf{low inter-cluster correlation}.
Second, we want an overall \textbf{low number of clusters}.
Intuitively, when looking at MAAD-Face, a high number of clusters indicates the retention of numerous correlated attributes that were not yet assigned to their ``suitable'' cluster.
Third and last, it is critical to achieve a \textbf{high retention of (genuine) samples}.
Undoubtedly, the more samples are retained for a specific cluster, the more expressive the results produced by any given analysis executed on the respective cluster are.
However, it is not sufficient to solely take into account the global number of retained samples.
Since we assess the effects of those attributes on Male and Female subgroups separately, it is critical to ensure that for each subgroup, the sampling requirements as per \cref{subsubsec:sampling} are met.
This is not trivial since, e.g., beards intuitively correlate more strongly with the male than with the female subgroup for biological reasons.
Consequently, we only view a clustering as valid if there is at least one label assignment to each cluster such that there are enough retained samples for each subgroup.

We evaluate the proposed decorrelation method under the stated criteria.
\begin{figure}[t]
	\centering
	\includegraphics[width=\linewidth]{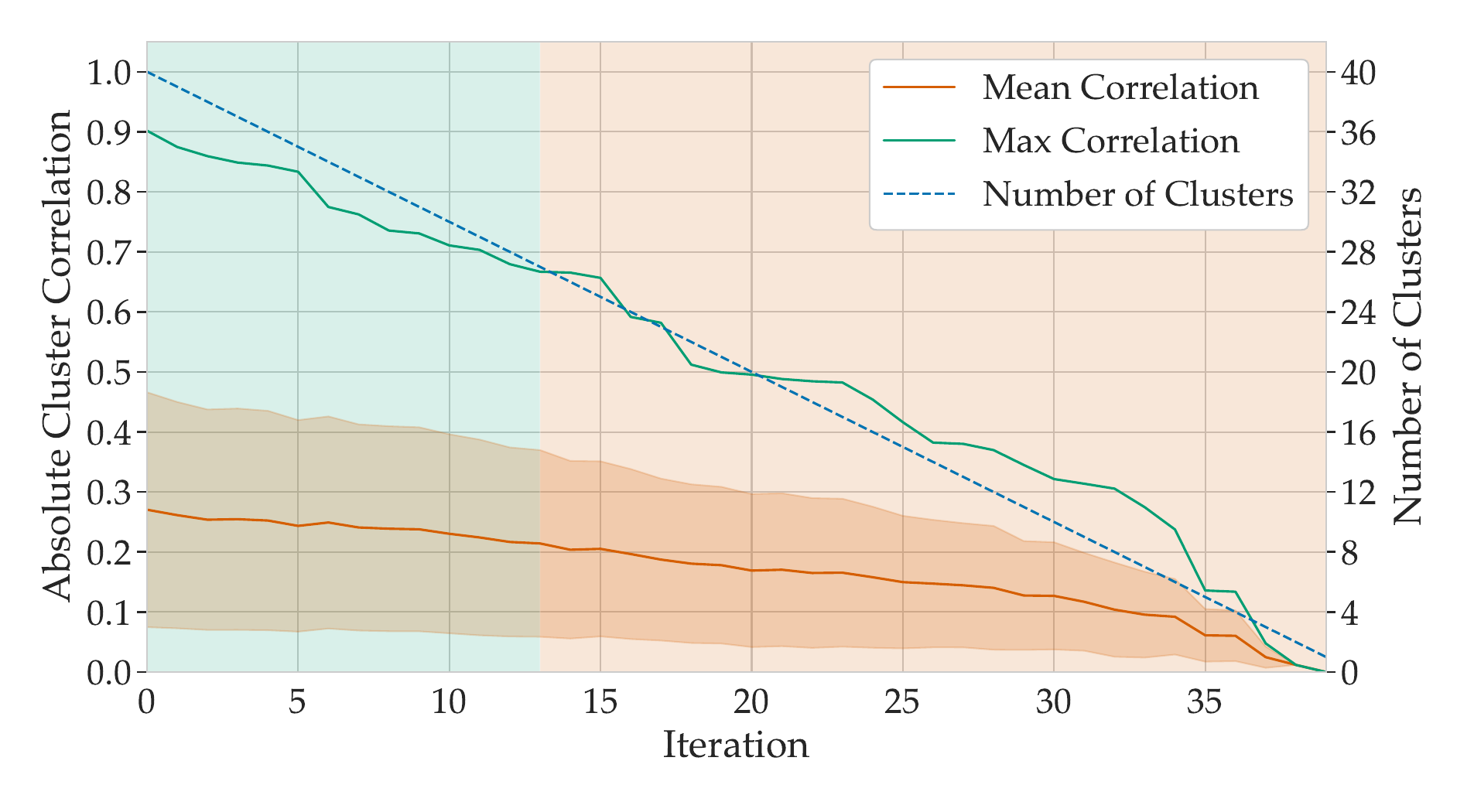}
	\caption{\textbf{Evolution of critical clustering metrics over progressing decorrelation} - The chosen measurements reflect the efficacy of the devised algorithm in reducing the correlation of MAAD-Face's non-demographic attributes. The iteration range reflects the clustering intensity, ranging from \num{0} (no clustering) to \num{39} (all attributes in one cluster). The number of clusters, mean and maximum correlation decrease linearly. For an optimal clustering w.r.t. low inter-cluster correlation and low number of clusters, it is therefore sensible to choose as late of an iteration as possible. To also accommodate the sampling requirements into these criteria, we conceive iteration \num{13} as optimal, since these requirements cannot be fulfilled thereafter (orange background).}
	\label{fig:clustering-evaluation-gender}
\end{figure}
\cref{fig:clustering-evaluation-gender} provides a visual representation of the key metrics' evolution over the algorithm's iterative progression.
As expected, a linear descent of the number of clusters w.r.t. the iterations can be witnessed.
This is intuitive since in each iteration, one cluster, which can either be a standalone attribute or a group thereof, is merged with another.
Moreover, we observe a steadily decreasing mean and maximum inter-cluster correlation.
From that behaviour, we can draw the interim conclusion that regarding the first two criteria, the optimal result seems to converge with progressing iterations.
However, we have not yet accounted for the retention of samples and associated sampling requirements, the third criterion.
Its evolution over the course of the iterations is indicated through the shading of the background.
The value space here is binary as the requirements can either be met (green) or not met (orange).
We see that for the first \num{13} iterations the requirements can be fulfilled, but not anytime thereafter.

Based on those observations, we can infer that the clustering algorithm should terminate after iteration \num{13}.
This conclusion is based on the observation that this iteration is the only one meeting the third criterion and, due to the converging nature of the related metrics, also the other two criteria.
This leaves us with \num{27} clusters having a mean inter-cluster correlation of approx. \num{0.21} with a standard deviation of about \num{0.16}.
The maximum inter-cluster correlation is  around \num{0.67}. 
Making these results more accessible, we name each resulting attribute-cluster containing more than one attribute.
To this end, we aim to find the most suitable description, best representing the semantics of the attributes the respective cluster is comprised of.
Our findings are shown in \cref{tab:attribute-clusters}.

\begin{table}[t]
    \centering
    \caption{\textbf{Clusters resulting from decorrelation} - Reported in conjunction with author-chosen descriptive titles representing the respective cluster semantics. Note that this table only shows results for clustered attributes, i.e., clusters with a size of greater than or equal to two. All non-demographic attributes that are part of MAAD-Face but do not show up here were not clustered and remain unchanged. Clustered and remaining unchanged attributes will be used in all further analyses.}
    \begin{tabular}{ll}
        \toprule
        \textbf{Feminine} & \textbf{Generic Facial Hair} \\
        \midrule
        Rosy Cheeks & 5 o'Clock Shadow \\
        Wearing Earrings & Sideburns \\
        Wearing Lipstick & Not No Beard \\
        Heavy Makeup & \multirow{3}{*}{
            \begin{tabular}{l}
                \toprule
                \textbf{Corpulent} \\
                \midrule
                Chubby \\
                Double Chin \\
                \bottomrule
            \end{tabular}
        } \\
        Attractive & \\
        Wavy Hair & \\
        Arched Eyebrows & \\
        \midrule
        \textbf{Frontal Facial Hair} & \textbf{Rugged} \\
        \midrule
        Goatee & Big Nose \\
        Mustache & Bags Under Eyes \\
        \midrule
        \textbf{Cheerful} & \textbf{Eyewear} \\
        \midrule
        High Cheekbones & Eyeglasses \\
        Smiling & Not No Eyewear \\
        \bottomrule
    \end{tabular}
    \label{tab:attribute-clusters}
\end{table}

All attributes originally dictated by MAAD-Face that do not occur in this table were not clustered and thus remain unchanged.
They will still be taken into account in all further analyses, in conjunction with the newly found clusters.
The decorrelated nature of this newly constructed attribute set contributes to the expressiveness and generalizability of this work's results.

\subsection{Equalizing Individual Attributes}
We begin with evaluating the fairness and performance of ArcFace and FaceNet.
To this end, we probe them with samples that result from individually equalizing the clusters the decorrelated attribute set comprises.
This is the first step towards projecting gender bias in face recognition to the presence or absence of a subset of non-demographic facial attributes.
Unbalanced data distributions are accounted for by applying the custom sampling technique as presented in \cref{subsubsec:sampling}, and can therefore be neglected in all further analyses.

The results of these investigations are presented in \cref{tab:individual-effects}.
\begin{table*}[]
	\centering
	\caption{\textbf{Face recognition fairness and performance on test samples filtered for individual decorrelated attributes} - Reported in terms of FNMR for a fixed FMR of $10^{-3}$ as well as the fairness metric iGARBE. The number of genuine samples indicates the scarcity of samples annotated with the respective attribute-label pair. Labels represent the enforced presence (\textcolor{green}{\cmark}) or absence (\textcolor{red}{\xmark}) of the corresponding attributes within samples. Pairs for which not enough samples could be retrieved to conform to the sampling requirements were left out. Although some assignments lead to increased fairness (higher iGARBE values), these improvements come at the cost of system performance (higher FNMR) most times. Perfect fairness cannot be achieved by solely inspecting the results of equalizing individual attributes.} 
	\begin{tabular}{llrrrrr}
		\toprule
		 & & \multicolumn{2}{c}{\makecell{ArcFace}} & \multicolumn{2}{c}{\makecell{FaceNet}} & \\ 
		\cmidrule(lr){3 - 4}
		\cmidrule(lr){5 - 6}
		         Attribute &                     Label &  iGARBE  &  FNMR 	 &  iGARBE  &  FNMR    & Genuine Samples \\
		\midrule
		               & \textcolor{lightgray}{$\bullet$} & $0.9592$ & $0.0714$ & $0.9253$ & $0.3682$ & $12257651$ \\
		\midrule
                 Bangs &   \textcolor{red}{\xmark} & $0.9566$ & $0.0656$ & $0.9138$ & $0.3570$ &     $7381868$ \\
                 Bangs & \textcolor{green}{\cmark} & $0.8382$ & $0.0891$ & $0.9345$ & $0.4787$ &      $384735$ \\
                  Bald &   \textcolor{red}{\xmark} & $0.9618$ & $0.0725$ & $0.9428$ & $0.3859$ &    $12213782$ \\
              Feminine &   \textcolor{red}{\xmark} & $0.7050$ & $0.0725$ & $0.7975$ & $0.3795$ &       $25041$ \\
            Blond Hair &   \textcolor{red}{\xmark} & $0.9528$ & $0.0703$ & $0.9236$ & $0.3502$ &     $4980573$ \\
     Receding Hairline &   \textcolor{red}{\xmark} & $0.9895$ & $0.0788$ & $0.9951$ & $0.4399$ &     $4525040$ \\
     Receding Hairline & \textcolor{green}{\cmark} & $0.7875$ & $0.0408$ & $0.8082$ & $0.2593$ &      $101650$ \\
            Brown Eyes &   \textcolor{red}{\xmark} & $0.7970$ & $0.0508$ & $0.8083$ & $0.3686$ &      $288447$ \\
            Brown Eyes & \textcolor{green}{\cmark} & $0.9355$ & $0.0826$ & $0.9190$ & $0.4154$ &     $1508420$ \\
       Wearing Necktie &   \textcolor{red}{\xmark} & $0.9687$ & $0.0813$ & $0.9593$ & $0.4471$ &     $5075688$ \\
             Corpulent &   \textcolor{red}{\xmark} & $0.9740$ & $0.0738$ & $0.9815$ & $0.4336$ &     $6978086$ \\
Fully Visible Forehead &   \textcolor{red}{\xmark} & $0.9622$ & $0.0916$ & $0.9843$ & $0.4557$ &     $1785040$ \\
Fully Visible Forehead & \textcolor{green}{\cmark} & $0.9205$ & $0.0502$ & $0.8641$ & $0.3159$ &     $3741913$ \\
            Black Hair &   \textcolor{red}{\xmark} & $0.9233$ & $0.0617$ & $0.8884$ & $0.3657$ &     $6613258$ \\
            Black Hair & \textcolor{green}{\cmark} & $0.9078$ & $0.0911$ & $0.9486$ & $0.4430$ &      $961940$ \\
          Mouth Closed &   \textcolor{red}{\xmark} & $0.9756$ & $0.0714$ & $0.9445$ & $0.4899$ &      $814259$ \\
           Wearing Hat &   \textcolor{red}{\xmark} & $0.9398$ & $0.0632$ & $0.9043$ & $0.3475$ &    $10892163$ \\
           Wearing Hat & \textcolor{green}{\cmark} & $0.9000$ & $0.1198$ & $0.9724$ & $0.5762$ &      $115686$ \\
                Rugged &   \textcolor{red}{\xmark} & $0.8543$ & $0.0753$ & $0.8913$ & $0.4770$ &      $210814$ \\
               Eyewear &   \textcolor{red}{\xmark} & $0.9154$ & $0.0609$ & $0.9166$ & $0.3370$ &     $9194376$ \\
   Obstructed Forehead &   \textcolor{red}{\xmark} & $0.9246$ & $0.0580$ & $0.8990$ & $0.3569$ &     $7822935$ \\
   Obstructed Forehead & \textcolor{green}{\cmark} & $0.9020$ & $0.1194$ & $0.9543$ & $0.6195$ &      $116882$ \\
           Pointy Nose &   \textcolor{red}{\xmark} & $0.8917$ & $0.0927$ & $0.9084$ & $0.4069$ &      $799108$ \\
           Pointy Nose & \textcolor{green}{\cmark} & $0.8641$ & $0.0592$ & $0.8582$ & $0.3830$ &     $4163088$ \\
              Cheerful &   \textcolor{red}{\xmark} & $0.9069$ & $0.0900$ & $0.9159$ & $0.4831$ &      $490084$ \\
              Cheerful & \textcolor{green}{\cmark} & $0.8891$ & $0.0522$ & $0.7307$ & $0.3439$ &       $23270$ \\
   Frontal Facial Hair &   \textcolor{red}{\xmark} & $0.9697$ & $0.0747$ & $0.9512$ & $0.3992$ &     $8838416$ \\
              Big Lips &   \textcolor{red}{\xmark} & $0.9241$ & $0.0621$ & $0.9342$ & $0.3400$ &     $1135456$ \\
              Big Lips & \textcolor{green}{\cmark} & $0.9399$ & $0.0769$ & $0.9883$ & $0.4361$ &     $2133293$ \\
            Shiny Skin &   \textcolor{red}{\xmark} & $0.9157$ & $0.0684$ & $0.9562$ & $0.4037$ &      $509064$ \\
            Shiny Skin & \textcolor{green}{\cmark} & $0.9320$ & $0.0606$ & $0.9083$ & $0.3963$ &      $593781$ \\
        Bushy Eyebrows &   \textcolor{red}{\xmark} & $0.9589$ & $0.0740$ & $0.9641$ & $0.4161$ &     $5259417$ \\
        Bushy Eyebrows & \textcolor{green}{\cmark} & $0.9207$ & $0.0545$ & $0.9794$ & $0.2957$ &       $57087$ \\
   Generic Facial Hair &   \textcolor{red}{\xmark} & $0.9707$ & $0.0793$ & $0.9874$ & $0.4218$ &     $1935348$ \\
             Gray Hair &   \textcolor{red}{\xmark} & $0.9718$ & $0.0743$ & $0.9490$ & $0.3905$ &    $10697373$ \\
            Brown Hair &   \textcolor{red}{\xmark} & $0.9529$ & $0.0692$ & $0.8720$ & $0.3616$ &     $1542963$ \\
            Brown Hair & \textcolor{green}{\cmark} & $0.9961$ & $0.0585$ & $0.9632$ & $0.4302$ &     $1714373$ \\
            Round Face &   \textcolor{red}{\xmark} & $0.9195$ & $0.0497$ & $0.8745$ & $0.3136$ &     $6461864$ \\
             Oval Face &   \textcolor{red}{\xmark} & $0.8729$ & $0.0929$ & $0.8928$ & $0.4030$ &      $427890$ \\
           Square Face &   \textcolor{red}{\xmark} & $0.9272$ & $0.0795$ & $0.9807$ & $0.4316$ &     $1011924$ \\
           Square Face & \textcolor{green}{\cmark} & $0.9826$ & $0.0600$ & $0.9809$ & $0.3148$ &       $91328$ \\
		\bottomrule
	\end{tabular}
	\label{tab:individual-effects}
\end{table*}
For each attribute, we provide results for ArcFace and FaceNet in terms of iGARBE and overall FNMR, as well as the number of genuine samples.
As can be observed, we retain enough samples to inspect both positively (\textcolor{green}{\cmark}) and negatively (\textcolor{red}{\xmark}) labeled assignments for \textit{Bangs}, \textit{Receding Hairline}, \textit{Brown Eyes}, \textit{Fully Visible Forehead}, \textit{Black Hair}, \textit{Wearing Hat}, \textit{Obstructed Forehead}, \textit{Pointy Nose}, \textit{Cheerful}, \textit{Big Lips}, \textit{Shiny Skin}, \textit{Bushy Eyebrows}, \textit{Brown Hair}, and \textit{Square Face}.
However, there exist attributes for which only negative but no positive label assignments are analyzed.
This is due to the omitted assignments not conforming with the sampling requirements we enforce to maintain high quality, highly expressive results.
Hence, we exclude these in our evaluation, retaining a total of \num{41} results.

For ArcFace, \num{11} attribute cluster assignments exceed baseline fairness.
These correspond to the absence of \textit{Generic Facial Hair} and \textit{Frontal Facial Hair}, not having the \textit{Mouth Closed}, not being \textit{Bald} or \textit{Corpulent}, not having \textit{Gray Hair} or a \textit{Fully Visible Forehead}, not \textit{Wearing Necktie}, having a \textit{Square Face}, and especially not having a \textit{Receding Hairline} as well as not having \textit{Brown Hair}.
The latter is particularly interesting, leading to an increased iGARBE fairness score of \num{0.9984} and a decreased FNMR of \num{0.0585} (corresponding to a system performance improvement of \qty{18}{\percent}) when probed on ArcFace.
With a retained \num{1.7} million genuine samples, the results are highly significant.
In nearly all other cases, an improved system fairness leads to a decreased system performance.
Crucially, we mitigate the probability of underlying correlations to be a factor in these outcomes due to the decorrelated nature of the attribute clusters.

Evaluating the results for the samples probed on FaceNet, it can be seen that \num{21} attribute cluster assignments exceed baseline fairness.
The best one of those corresponds to the absence of a \textit{Receding Hairline}.
It leads to an increased iGARBE score of \num{0.9951} but also to an increased FNMR of \num{0.4218} (corresponding to an $\approx$\qty{19}{\percent} system performance deterioration).
Approximately \num{4.5} million matching genuine samples are retained.
Interestingly, this particular assignment also is the second best performing attribute for ArcFace.
Moreover, the corresponding observations made match those asserted with ArcFace as well.
Most times, the equalization of individual attribute clusters across genders shows an inverse relationship between system fairness and system performance.
The only significant outlier for FaceNet in this respect, showing high iGARBE scores as well as leading to significantly improved system performance, results from probing the system with samples labeled as having \textit{Bushy Eyebrows}.
Doing so yields an iGARBE score of \num{0.9794} and an FNMR of \num{0.2957}.
However, at only \num{57} thousand retained genuine samples, the expressiveness of these results is reduced compared to those assignments with a genuine sample retention in the millions as supported by the BioQuake 1\% rule of \textcite{Fallahi2025OnTR}. 

For both FRS, the remaining results match these findings and show no other interesting features.
As can be deduced, filtering only for individual attribute clusters falls short of baseline fairness in most cases.
If it is exceeded, then this fairness-increasing characteristic can only be achieved at the expense of system performance (aside for some few outliers).
However, even these outliers do not systematically converge to perfect fairness scores of \num{1}.
These observations reinforce both the plausibility and the necessity of focusing on an approach that combines multiple attribute clusters to uncover which \emph{combinations} affect gender bias the most, ideally without deteriorating false-negative error rates.
In doing so and, thus, by applying this work's proposed methodology, we can therefore be more productive w.r.t. the research objective at hand.

\subsection{Equalizing Attribute Combinations}
\subsubsection{Performance}
We proceed to inspect how combinations of attribute cluster assignments affect system fairness and performance.
Therefore, we apply the methodology as proposed in \cref{subsec:forming-combinations}.
To this end, we will search for optimal combinations w.r.t. system fairness separately for ArcFace and FaceNet to account for the varying topology of the optimization problem.
Such differences are most likely caused by the different loss functions used in training the FRS.
Nonetheless, we will evaluate the corresponding optimal combinations for the respective other system as well.
We report our results in \cref{tab:combined}.

\begin{table*}[]
	\centering
	\caption{\textbf{Combinations of decorrelated attributes achieving ten highest iGARBE scores} - Reported in terms of FNMR for a fixed FMR of $10^{-3}$, the fairness metric iGARBE and the contextualized fairness metric CoFair. Again, the number of genuine samples indicates the scarcity of samples annotated with the given assignment combination. Labels represent the enforced presence (\textcolor{green}{\cmark}) or absence (\textcolor{red}{\xmark}) of the corresponding attributes within samples, or no enforcement (\textcolor{lightgray}{$\bullet$}). As the combinations found by the proposed unsupervised approach are specific per FRS, we report them separately for ArcFace and FaceNet in Tables \ref{tab:combined-arcface} and \ref{tab:combined-facenet}. To underline that the resulting combinations also have a positive impact on fairness for the respective other FRS, we report iGARBE and CoFair values for them as well.}
	\begin{subtable}[t]{\textwidth}
		\centering
		\caption{Top 10 gender-balanced (fair) attribute combinations based on ArcFace}
		\begin{adjustbox}{max width=\textwidth}
			\begin{tabular}{ccccccccccccrrrrrrrr}
				\toprule
  				\multirow[b]{7}{*}[-9pt]{\rot{Receding Hairline}} & \multirow[b]{7}{*}[-9pt]{\rot{Bushy Eyebrows}} & \multirow[b]{7}{*}[-10pt]{\rot{Blond Hair}} & \multirow[b]{7}{*}[-8pt]{\rot{Eyewear}} & \multirow[b]{7}{*}[-8pt]{\rot{Frontal Facial Hair}} & \multirow[b]{7}{*}[-8pt]{\rot{Gray Hair}} & \multirow[b]{7}{*}[-8pt]{\rot{Bangs}} & \multirow[b]{7}{*}[-8pt]{\rot{Corpulent}} & \multirow[b]{7}{*}[-8pt]{\rot{Wearing Hat}} & \multirow[b]{7}{*}[-8pt]{\rot{Wearing Necktie}} & \multirow[b]{7}{*}[-10pt]{\rot{Bald}} & \multirow[b]{7}{*}[-10pt]{\rot{Black Hair}} & & & & & & & & \\
  				& & & & & & & & & & & & & & & & & & \\
  				& & & & & & & & & & & & & & & & & & \\
  				& & & & & & & & & & & & & & & & & & \\
  				& & & & & & & & & & & & & & & & & & \\
  				& & & & & & & & & & & & \multicolumn{3}{c}{\makecell{FNMR - ArcFace}} & \multicolumn{2}{c}{\makecell{iGARBE}} & \multicolumn{2}{c}{\makecell{$\mathrm{CoFair}$}} & \multirow[b]{2}{*}[-6pt]{Gen. Samples} \\
				\cmidrule(lr){13 - 15} 
				\cmidrule(lr){16 - 17}
				\cmidrule(lr){18 - 19}
				  & & & & & & & & & & & & Male & Female & Total & ArcFace & FaceNet & ArcFace & FaceNet & \\
				\midrule
          \textcolor{lightgray}{$\bullet$} &                \textcolor{lightgray}{$\bullet$} &            \textcolor{lightgray}{$\bullet$} &         \textcolor{lightgray}{$\bullet$} &                   \textcolor{lightgray}{$\bullet$} &           \textcolor{lightgray}{$\bullet$} &       \textcolor{lightgray}{$\bullet$} &           \textcolor{lightgray}{$\bullet$} &             \textcolor{lightgray}{$\bullet$} &                 \textcolor{lightgray}{$\bullet$} &      \textcolor{lightgray}{$\bullet$} &            \textcolor{lightgray}{$\bullet$} & $0.0684$ & $0.0784$ & $0.0714$ & $0.9592$ & $0.9253$ & $0.7266$ & $0.6639$ & $12257651$ \\
        \midrule
          \textcolor{lightgray}{$\bullet$} &                \textcolor{lightgray}{$\bullet$} &                     \textcolor{red}{\xmark} &         \textcolor{lightgray}{$\bullet$} &                   \textcolor{lightgray}{$\bullet$} &                    \textcolor{red}{\xmark} &       \textcolor{lightgray}{$\bullet$} &                    \textcolor{red}{\xmark} &             \textcolor{lightgray}{$\bullet$} &                 \textcolor{lightgray}{$\bullet$} &      \textcolor{lightgray}{$\bullet$} &                     \textcolor{red}{\xmark} & $0.0594$ & $0.0593$ & $0.0590$ & $0.9987$ & $0.9664$ & $0.9975$ & $0.9134$ &  $1170030$ \\
          \textcolor{lightgray}{$\bullet$} &                \textcolor{lightgray}{$\bullet$} &            \textcolor{lightgray}{$\bullet$} &                  \textcolor{red}{\xmark} &                   \textcolor{lightgray}{$\bullet$} &           \textcolor{lightgray}{$\bullet$} &       \textcolor{lightgray}{$\bullet$} &           \textcolor{lightgray}{$\bullet$} &             \textcolor{lightgray}{$\bullet$} &                          \textcolor{red}{\xmark} &               \textcolor{red}{\xmark} &            \textcolor{lightgray}{$\bullet$} & $0.0710$ & $0.0714$ & $0.0715$ & $0.9983$ & $0.9842$ & $0.9968$ & $0.9723$ &  $2882629$ \\
          \textcolor{lightgray}{$\bullet$} &                         \textcolor{red}{\xmark} &                     \textcolor{red}{\xmark} &         \textcolor{lightgray}{$\bullet$} &                            \textcolor{red}{\xmark} &                    \textcolor{red}{\xmark} &       \textcolor{lightgray}{$\bullet$} &           \textcolor{lightgray}{$\bullet$} &             \textcolor{lightgray}{$\bullet$} &                 \textcolor{lightgray}{$\bullet$} &      \textcolor{lightgray}{$\bullet$} &            \textcolor{lightgray}{$\bullet$} & $0.0782$ & $0.0777$ & $0.0781$ & $0.9982$ & $0.9684$ & $0.9968$ & $0.9219$ &  $2242832$ \\
          \textcolor{lightgray}{$\bullet$} &                \textcolor{lightgray}{$\bullet$} &                     \textcolor{red}{\xmark} &         \textcolor{lightgray}{$\bullet$} &                            \textcolor{red}{\xmark} &           \textcolor{lightgray}{$\bullet$} &       \textcolor{lightgray}{$\bullet$} &           \textcolor{lightgray}{$\bullet$} &             \textcolor{lightgray}{$\bullet$} &                 \textcolor{lightgray}{$\bullet$} &      \textcolor{lightgray}{$\bullet$} &                     \textcolor{red}{\xmark} & $0.0591$ & $0.0594$ & $0.0585$ & $0.9979$ & $0.9660$ & $0.9962$ & $0.9122$ &  $1363650$ \\
          \textcolor{lightgray}{$\bullet$} &                \textcolor{lightgray}{$\bullet$} &                     \textcolor{red}{\xmark} &         \textcolor{lightgray}{$\bullet$} &                   \textcolor{lightgray}{$\bullet$} &           \textcolor{lightgray}{$\bullet$} &       \textcolor{lightgray}{$\bullet$} &           \textcolor{lightgray}{$\bullet$} &                      \textcolor{red}{\xmark} &                          \textcolor{red}{\xmark} &               \textcolor{red}{\xmark} &            \textcolor{lightgray}{$\bullet$} & $0.0721$ & $0.0726$ & $0.0724$ & $0.9979$ & $0.9665$ & $0.9962$ & $0.9147$ &  $2634757$ \\
                   \textcolor{red}{\xmark} &                \textcolor{lightgray}{$\bullet$} &            \textcolor{lightgray}{$\bullet$} &         \textcolor{lightgray}{$\bullet$} &                   \textcolor{lightgray}{$\bullet$} &                    \textcolor{red}{\xmark} &                \textcolor{red}{\xmark} &           \textcolor{lightgray}{$\bullet$} &             \textcolor{lightgray}{$\bullet$} &                 \textcolor{lightgray}{$\bullet$} &               \textcolor{red}{\xmark} &            \textcolor{lightgray}{$\bullet$} & $0.0725$ & $0.0719$ & $0.0724$ & $0.9978$ & $0.9378$ & $0.9955$ & $0.7538$ &  $2114894$ \\
          \textcolor{lightgray}{$\bullet$} &                         \textcolor{red}{\xmark} &            \textcolor{lightgray}{$\bullet$} &         \textcolor{lightgray}{$\bullet$} &                            \textcolor{red}{\xmark} &                    \textcolor{red}{\xmark} &       \textcolor{lightgray}{$\bullet$} &           \textcolor{lightgray}{$\bullet$} &             \textcolor{lightgray}{$\bullet$} &                 \textcolor{lightgray}{$\bullet$} &               \textcolor{red}{\xmark} &            \textcolor{lightgray}{$\bullet$} & $0.0764$ & $0.0770$ & $0.0775$ & $0.9977$ & $0.9491$ & $0.9955$ & $0.8267$ &  $2234224$ \\
                   \textcolor{red}{\xmark} &                \textcolor{lightgray}{$\bullet$} &                     \textcolor{red}{\xmark} &         \textcolor{lightgray}{$\bullet$} &                   \textcolor{lightgray}{$\bullet$} &           \textcolor{lightgray}{$\bullet$} &       \textcolor{lightgray}{$\bullet$} &           \textcolor{lightgray}{$\bullet$} &             \textcolor{lightgray}{$\bullet$} &                 \textcolor{lightgray}{$\bullet$} &               \textcolor{red}{\xmark} &            \textcolor{lightgray}{$\bullet$} & $0.0821$ & $0.0814$ & $0.0816$ & $0.9976$ & $0.9641$ & $0.9955$ & $0.9044$ &  $3827023$ \\
          \textcolor{lightgray}{$\bullet$} &                \textcolor{lightgray}{$\bullet$} &            \textcolor{lightgray}{$\bullet$} &         \textcolor{lightgray}{$\bullet$} &                   \textcolor{lightgray}{$\bullet$} &           \textcolor{lightgray}{$\bullet$} &       \textcolor{lightgray}{$\bullet$} &                    \textcolor{red}{\xmark} &                      \textcolor{red}{\xmark} &                          \textcolor{red}{\xmark} &               \textcolor{red}{\xmark} &            \textcolor{lightgray}{$\bullet$} & $0.0699$ & $0.0705$ & $0.0705$ & $0.9974$ & $0.9798$ & $0.9948$ & $0.9600$ &  $2415037$ \\
          \textcolor{lightgray}{$\bullet$} &                \textcolor{lightgray}{$\bullet$} &            \textcolor{lightgray}{$\bullet$} &         \textcolor{lightgray}{$\bullet$} &                            \textcolor{red}{\xmark} &           \textcolor{lightgray}{$\bullet$} &                \textcolor{red}{\xmark} &                    \textcolor{red}{\xmark} &             \textcolor{lightgray}{$\bullet$} &                 \textcolor{lightgray}{$\bullet$} &               \textcolor{red}{\xmark} &            \textcolor{lightgray}{$\bullet$} & $0.0691$ & $0.0697$ & $0.0700$ & $0.9974$ & $0.9628$ & $0.9948$ & $0.8990$ &  $2455854$ \\
				\bottomrule
			\end{tabular}
		\end{adjustbox}
		\label{tab:combined-arcface}
	\end{subtable}
	
	\vspace{5mm}
	
	\begin{subtable}[t]{\textwidth}
		\centering
		\caption{Top 10 gender-balanced (fair) attribute combinations based on FaceNet}
		\begin{adjustbox}{max width=\textwidth}
			\begin{tabular}{cccccccccccrrrrrrrr}
				\toprule
  				\multirow[b]{8}{*}[-8pt]{\rot{Wearing Hat}} & \multirow[b]{8}{*}[-9pt]{\rot{Corpulent}} & \multirow[b]{8}{*}[-9pt]{\rot{Gray Hair}} & \multirow[b]{8}{*}[-8pt]{\rot{Big Lips}} & \multirow[b]{8}{*}[-8pt]{\rot{Bushy Eyebrows}} & \multirow[b]{8}{*}[-9pt]{\rot{Blond Hair}} & \multirow[b]{8}{*}[-9pt]{\rot{Bald}} & \multirow[b]{8}{*}[-9pt]{\rot{Frontal Facial Hair}} & \multirow[b]{8}{*}[-10pt]{\rot{Obstructed Forehead}} & \multirow[b]{8}{*}[-8pt]{\rot{Bangs}} & \multirow[b]{8}{*}[-8pt]{\rot{Eyewear}} & & & & & & & \\
  				& & & & & & & & & & & & & & & & \\
  				& & & & & & & & & & & & & & & & \\
  				& & & & & & & & & & & & & & & & \\
  				& & & & & & & & & & & & & & & & \\
          		& & & & & & & & & & & & & & & & \\
  				& & & & & & & & & & & \multicolumn{3}{c}{\makecell{FNMR - FaceNet}} & \multicolumn{2}{c}{\makecell{iGARBE}} & \multicolumn{2}{c}{\makecell{$\mathrm{CoFair}$}} & \multirow[b]{2}{*}[-6pt]{Gen. Samples}\\
				\cmidrule(lr){12 - 14} 
				\cmidrule(lr){15 - 16}
				\cmidrule(lr){17 - 18}
				  & & & & & & & & & & & Male & Female & Total & FaceNet & ArcFace & FaceNet & ArcFace & \\
				\midrule
          \textcolor{lightgray}{$\bullet$} &           \textcolor{lightgray}{$\bullet$} &           \textcolor{lightgray}{$\bullet$} &          \textcolor{lightgray}{$\bullet$} &                \textcolor{lightgray}{$\bullet$} &            \textcolor{lightgray}{$\bullet$} &      \textcolor{lightgray}{$\bullet$} &                   \textcolor{lightgray}{$\bullet$} &                   \textcolor{lightgray}{$\bullet$} &       \textcolor{lightgray}{$\bullet$} &         \textcolor{lightgray}{$\bullet$} & $0.3332$ & $0.4268$ & $0.3682$ & $0.9253$ & $0.9592$ & $0.6639$ & $0.7266$ & $12345855$ \\
        \midrule
          \textcolor{lightgray}{$\bullet$} &                    \textcolor{red}{\xmark} &                    \textcolor{red}{\xmark} &          \textcolor{lightgray}{$\bullet$} &                \textcolor{lightgray}{$\bullet$} &            \textcolor{lightgray}{$\bullet$} &      \textcolor{lightgray}{$\bullet$} &                   \textcolor{lightgray}{$\bullet$} &                            \textcolor{red}{\xmark} &       \textcolor{lightgray}{$\bullet$} &                  \textcolor{red}{\xmark} & $0.4092$ & $0.4096$ & $0.4101$ & $0.9997$ & $0.9764$ & $0.9997$ & $0.8997$ &  $2652903$ \\
                   \textcolor{red}{\xmark} &                    \textcolor{red}{\xmark} &                    \textcolor{red}{\xmark} &          \textcolor{lightgray}{$\bullet$} &                \textcolor{lightgray}{$\bullet$} &            \textcolor{lightgray}{$\bullet$} &      \textcolor{lightgray}{$\bullet$} &                            \textcolor{red}{\xmark} &                   \textcolor{lightgray}{$\bullet$} &       \textcolor{lightgray}{$\bullet$} &         \textcolor{lightgray}{$\bullet$} & $0.4349$ & $0.4335$ & $0.4331$ & $0.9991$ & $0.9771$ & $0.9991$ & $0.9051$ &  $2672231$ \\
                   \textcolor{red}{\xmark} &                    \textcolor{red}{\xmark} &           \textcolor{lightgray}{$\bullet$} &          \textcolor{lightgray}{$\bullet$} &                \textcolor{lightgray}{$\bullet$} &            \textcolor{lightgray}{$\bullet$} &      \textcolor{lightgray}{$\bullet$} &                   \textcolor{lightgray}{$\bullet$} &                            \textcolor{red}{\xmark} &       \textcolor{lightgray}{$\bullet$} &                  \textcolor{red}{\xmark} & $0.4035$ & $0.4051$ & $0.4049$ & $0.9988$ & $0.9704$ & $0.9988$ & $0.8450$ &  $2805567$ \\
          \textcolor{lightgray}{$\bullet$} &           \textcolor{lightgray}{$\bullet$} &           \textcolor{lightgray}{$\bullet$} &          \textcolor{lightgray}{$\bullet$} &                         \textcolor{red}{\xmark} &            \textcolor{lightgray}{$\bullet$} &               \textcolor{red}{\xmark} &                            \textcolor{red}{\xmark} &                   \textcolor{lightgray}{$\bullet$} &       \textcolor{lightgray}{$\bullet$} &         \textcolor{lightgray}{$\bullet$} & $0.4308$ & $0.4318$ & $0.4306$ & $0.9986$ & $0.9266$ & $0.9988$ & $0.4298$ &  $2892688$ \\
          \textcolor{lightgray}{$\bullet$} &           \textcolor{lightgray}{$\bullet$} &           \textcolor{lightgray}{$\bullet$} &          \textcolor{lightgray}{$\bullet$} &                         \textcolor{red}{\xmark} &                     \textcolor{red}{\xmark} &               \textcolor{red}{\xmark} &                            \textcolor{red}{\xmark} &                   \textcolor{lightgray}{$\bullet$} &       \textcolor{lightgray}{$\bullet$} &         \textcolor{lightgray}{$\bullet$} & $0.4353$ & $0.4374$ & $0.4369$ & $0.9986$ & $0.9782$ & $0.9988$ & $0.9129$ &  $2576408$ \\
          \textcolor{lightgray}{$\bullet$} &           \textcolor{lightgray}{$\bullet$} &                    \textcolor{red}{\xmark} &          \textcolor{lightgray}{$\bullet$} &                         \textcolor{red}{\xmark} &            \textcolor{lightgray}{$\bullet$} &               \textcolor{red}{\xmark} &                   \textcolor{lightgray}{$\bullet$} &                   \textcolor{lightgray}{$\bullet$} &                \textcolor{red}{\xmark} &         \textcolor{lightgray}{$\bullet$} & $0.4409$ & $0.4432$ & $0.4415$ & $0.9985$ & $0.9442$ & $0.9985$ & $0.5868$ &  $2355399$ \\
          \textcolor{lightgray}{$\bullet$} &                    \textcolor{red}{\xmark} &           \textcolor{lightgray}{$\bullet$} &          \textcolor{lightgray}{$\bullet$} &                \textcolor{lightgray}{$\bullet$} &            \textcolor{lightgray}{$\bullet$} &      \textcolor{lightgray}{$\bullet$} &                   \textcolor{lightgray}{$\bullet$} &                            \textcolor{red}{\xmark} &                \textcolor{red}{\xmark} &                  \textcolor{red}{\xmark} & $0.4092$ & $0.4113$ & $0.4102$ & $0.9984$ & $0.9642$ & $0.9985$ & $0.7822$ &  $2723159$ \\
                   \textcolor{red}{\xmark} &           \textcolor{lightgray}{$\bullet$} &           \textcolor{lightgray}{$\bullet$} &                   \textcolor{red}{\xmark} &                         \textcolor{red}{\xmark} &            \textcolor{lightgray}{$\bullet$} &               \textcolor{red}{\xmark} &                   \textcolor{lightgray}{$\bullet$} &                   \textcolor{lightgray}{$\bullet$} &       \textcolor{lightgray}{$\bullet$} &         \textcolor{lightgray}{$\bullet$} & $0.3866$ & $0.3889$ & $0.3853$ & $0.9978$ & $0.9858$ & $0.9978$ & $0.9590$ &   $927385$ \\
          \textcolor{lightgray}{$\bullet$} &                    \textcolor{red}{\xmark} &           \textcolor{lightgray}{$\bullet$} &          \textcolor{lightgray}{$\bullet$} &                \textcolor{lightgray}{$\bullet$} &            \textcolor{lightgray}{$\bullet$} &               \textcolor{red}{\xmark} &                   \textcolor{lightgray}{$\bullet$} &                            \textcolor{red}{\xmark} &       \textcolor{lightgray}{$\bullet$} &                  \textcolor{red}{\xmark} & $0.4133$ & $0.4098$ & $0.4098$ & $0.9974$ & $0.9768$ & $0.9975$ & $0.9024$ &  $2555046$ \\
          \textcolor{lightgray}{$\bullet$} &           \textcolor{lightgray}{$\bullet$} &                    \textcolor{red}{\xmark} &          \textcolor{lightgray}{$\bullet$} &                         \textcolor{red}{\xmark} &            \textcolor{lightgray}{$\bullet$} &      \textcolor{lightgray}{$\bullet$} &                            \textcolor{red}{\xmark} &                   \textcolor{lightgray}{$\bullet$} &       \textcolor{lightgray}{$\bullet$} &                  \textcolor{red}{\xmark} & $0.3997$ & $0.3960$ & $0.3955$ & $0.9972$ & $0.9818$ & $0.9972$ & $0.9379$ &  $1522274$ \\
				\bottomrule
			\end{tabular}
		\end{adjustbox}
		\label{tab:combined-facenet}
	\end{subtable}
	\label{tab:combined}
\end{table*}

The results of applying the proposed approach to find optimal results for ArcFace are presented in \cref{tab:combined-arcface}.
All assignment combinations clearly exceed baseline fairness with values ranging from \num{0.9974} to \num{0.9987}.
Considering that a maximum possible iGARBE score of \num{1} indicates perfect fairness, these high scores are especially noteworthy.
Most of the assignments responsible for these results are related to (facial) hair.
\textit{Receding Hairline}, \textit{Bushy Eyebrows}, \textit{Blond Hair}, \textit{Frontal Facial Hair}, \textit{Gray Hair}, \textit{Bangs}, \textit{Bald}, and \textit{Black Hair} all fall under that category.
The second most prominent attribute clusters can be categorized as ``occluding accessories'',  namely \textit{Eyewear}, \textit{Wearing Hat}, and \textit{Wearing Necktie}.
\textit{Corpulent} is the only outlier.
Given this observation, it can be inferred that for ArcFace, a lack of facial hair and the absence of certain hair colors but the overall presence of hair, is beneficial for equal recognition performance across gender.
Paired with the absence of specific occluding accessories in some cases, we can deduct that the masking of specific facial areas or the lack thereof is an important factor for differential outcome across genders for ArcFace.
The sporadic occurrence or absence of \textit{Corpulent} throughout the results can be related to this inference using the common notion that an equal amount of face representation is paramount for cross-gender fairness, which this cluster allows more control over.
Otherwise, there seems to be no clear pattern of which combinations occur regularly, i.e., which attribute clusters are often paired with others or not.

The overall system performance increases for four out of ten assignment combinations.
In these cases, the average decrease in FNMR is about \qty{10}{\percent}.
Assignments involved only in such situations are to \textit{Corpulent} and \textit{Black Hair}.
In the remaining instances, the total FNMR increases slightly, on average by about \qty{5.5}{\percent}.
The recognition errors for Male-labeled and Female-labeled samples are similar.
Consequently, the inferences made above also apply for the gender-specific cases.
It should only be noted that the relative improvements differ slightly due to a different baseline.
The number of retained genuine samples lies within a range of around \num{1170000} and \num{3820000}, reinforcing the representativeness of the results.

\begin{figure}[t]
	\centering
	\includegraphics[width=\linewidth]{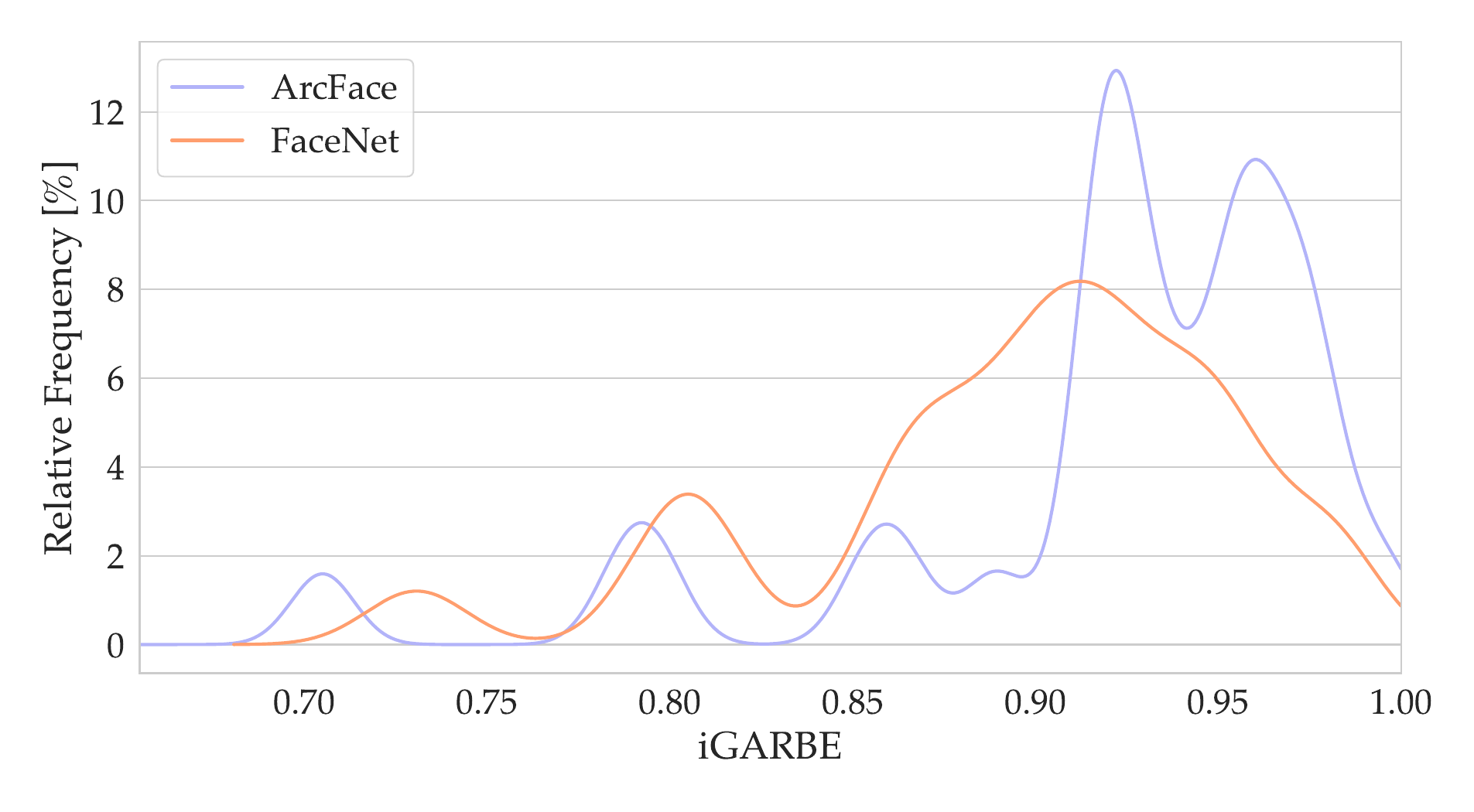}
	\caption{\textbf{Fairness iGARBE distributions for both FRS} - The estimations were performed using a KDE with Gaussian kernels with a bandwidth determined using Scott's rule. These distributions are the basis for CoFair. 
    }
	\label{fig:combined-fairness-dist-kde}
\end{figure}

To reason about the values of $\mathrm{CoFair}$, we first need to elaborate on their calculation.
As explained in \cref{subsubsec:metric-contextualization}, the purpose of $\mathrm{CoFair}$ is the contextualization of iGARBE fairness scores relative to the underlying distribution characteristics of the respective FRS.
Therefore, we need to estimate the probability density of iGARBE scores to calculate the cumulative distribution function.
To this end, we perform a kernel density estimation (KDE) with Gaussian kernels for a bandwidth estimated using Scott's rule \cite{scottMultivariateDensityEstimation1992}.
We multiply the resulting bandwidth with an additional adjusting factor of \num{0.5} to prevent oversmoothing.
For the estimation, we use the samples from \cref{tab:individual-effects}.
We specifically use only those samples whose FNMR is better than or at max \qty{10}{\percent} worse than the corresponding baseline FNMR.
In doing so, we specifically exclude samples with an unrepresentative performance that would therefore not contribute to an usable, realistic estimation.
The resulting probability density is shown in \cref{fig:combined-fairness-dist-kde}.

We now shift the focus on the results of $\mathrm{CoFair}$ for each FRS.
For ArcFace, results ranging from \num{0.9948} to \num{0.9975} at a baseline of \num{0.7266} can be observed in this respect.
Recall that a value of $\mathrm{CoFair}=0.9948$ indicates that the respective combination achieves an iGARBE fairness score that is higher than that of \qty{99.48}{\percent} of the single attributes on the same FRS.
Therefore, the optimization approach works as expected.
At the same time, these results are also to be expected, since the underlying iGARBE scores already approach values resembling nearly perfect fairness (i.e., $\mathrm{iGARBE}=1$).
However, for the given values, results of $\mathrm{CoFair}=1$ are not reached because of the positive response of ArcFace to people with \textit{Brown Hair} in terms of fairness.
When inspecting the impact of the ArcFace-focussed combinations on the fairness of FaceNet, contextualizing iGARBE scores offers more insight into the effectiveness of the presented approach.
The scores themselves range from \num{0.9378} to \num{0.9842} at a baseline of \num{0.9253}.
Although these improvements might seem minor at first, contextualizing enables a more comprehensive analysis.
With a baseline of $\mathrm{CoFair}=0.6639$, the results range from \num{0.7538} to \num{0.9723}, with \num{7} out of \num{10} values exceeding \num{0.9}.
These results underline the capability of the proposed approach in finding facial characteristics positively impacting gender fairness, ubiquitously.

FaceNet-specific results, as shown in \cref{tab:combined-facenet}, are semantically similar to those of ArcFace.
Again, the found attribute cluster assignment combinations exceed the baseline fairness score of \num{0.9253} by far.
They range from \num{0.9972} to \num{0.9997}, thus approaching the perfect score of \num{1}, as was the case for ArcFace.
Most of the clustered attributes associated with these results as part of combinations were already seen and discussed in the ArcFace-focused analysis, namely all but \textit{Big Lips} and \textit{Obstructed Forehead}.
Consequences observed and inferences made based on them and their negative label assignments largely parallel those for ArcFace.
As for the specific absence of \textit{Obstructed Forehead}, it can be inferred that accessories or hairstyles occluding the forehead in particular seem to be a dictating factor in affecting recognition error differentials.
Negatively labeled \textit{Big Lips} also seems to play a role in fostering the gender gap.
This is likely to be due to its connection to controlling the appearance of the mouth region of the face, which therefore also appears to be vital for recognition performance.
At the same time, since such an assignment only occurs in one of the top ten results, its significance is reduced compared to the other attributes.
Again, no clear pattern emerges regarding which attributes are combined in which specific scenario.
Genuine sample retention is again predominantly in the millions, ranging from about \num{920000} to \num{2890000}.
The total FNMR also increases by approx. \qty{17}{\percent} compared to the baseline, thereby decreasing overall recognition performance.

To better understand the iGARBE scores w.r.t. the underlying distribution characteristics, we again compute the contextualized fairness  $\mathrm{CoFair}$.
For FaceNet, this results in values between \num{0.9972} and \num{0.9997}.
At a baseline of \num{0.6639}, these outcomes reinforce the capabilities of the presented approach.
Once more, these values are to be expected since our results approach perfect fairness, i.e., iGARBE scores of \num{1}.
The iGARBE scores for ArcFace in the ArcFace-optimized approach are comparable to those for FaceNet in the FaceNet-focused analysis.
However, the results of $\mathrm{CoFair}$ for the latter are slightly higher regardless.
This is due to there not being as high-scored peak in the probability distribution as was the case for ArcFace due to the impact of \textit{Brown Hair}, as can be seen in \cref{fig:combined-fairness-dist-kde}.
Shifting the attention to the performance of ArcFace on combinations created with a focus on FaceNet, the respective iGARBE scores are in between \num{0.9266} and \num{0.9858} at a baseline of \num{0.9592}.
The corresponding values of $\mathrm{CoFair}$ range from \num{0.4298} to \num{0.9590} at a baseline of \num{0.7266}.
\num{5} out of \num{10} combinations exceed a value of \num{0.9}, again proving the capability of the presented approach w.r.t. unveiling facial characteristics positively affecting gender fairness in FRS.

\begin{figure}[!h]
	\begin{subfigure}{.5\textwidth}
		\centering
		\includegraphics[width=\linewidth]{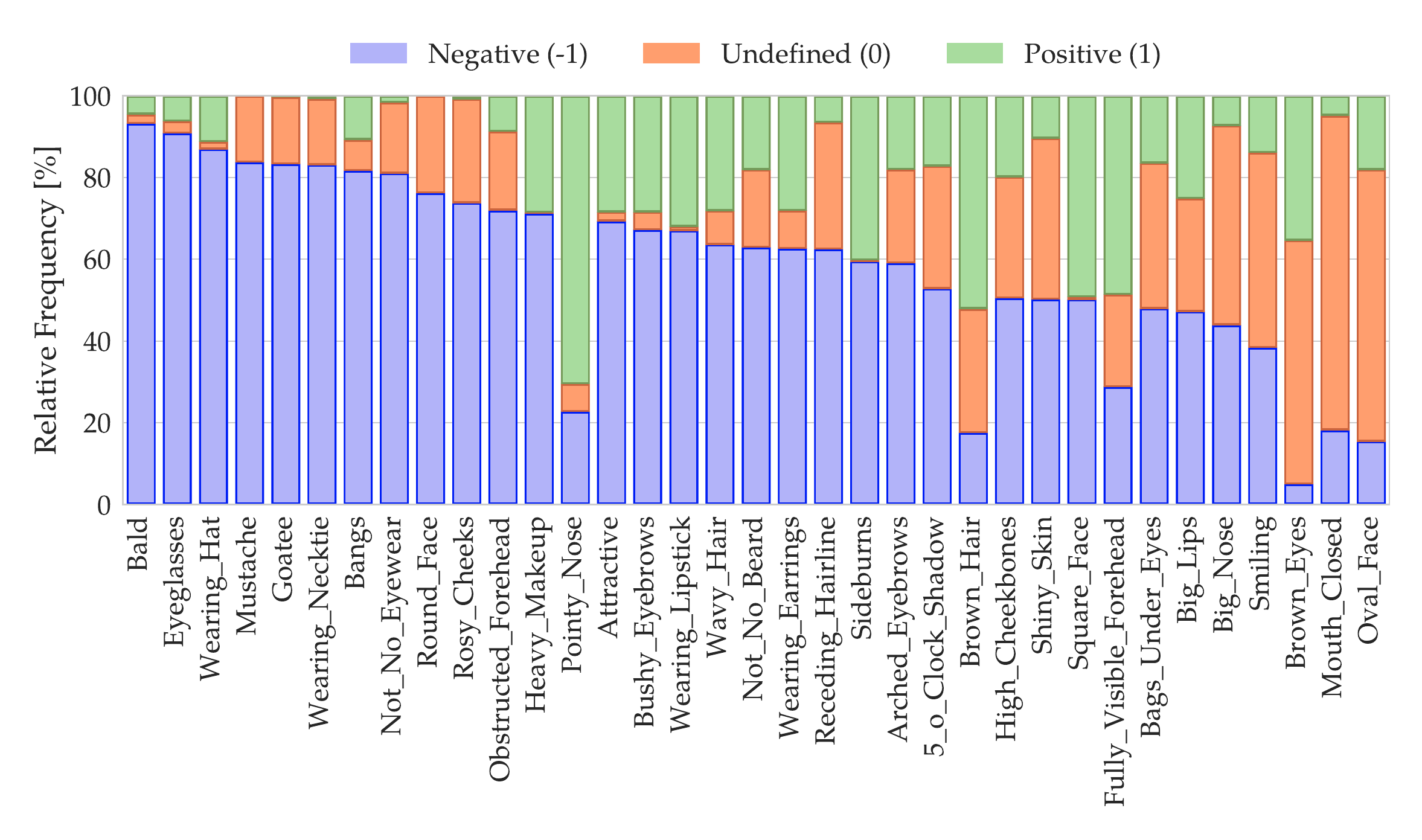}
		\caption{ArcFace}
		\label{fig:assignment-dist-arcface-clustered}
	\end{subfigure}%
    \vspace{-0.3em}
	\begin{subfigure}{.5\textwidth}
		\centering
		\includegraphics[width=\linewidth]{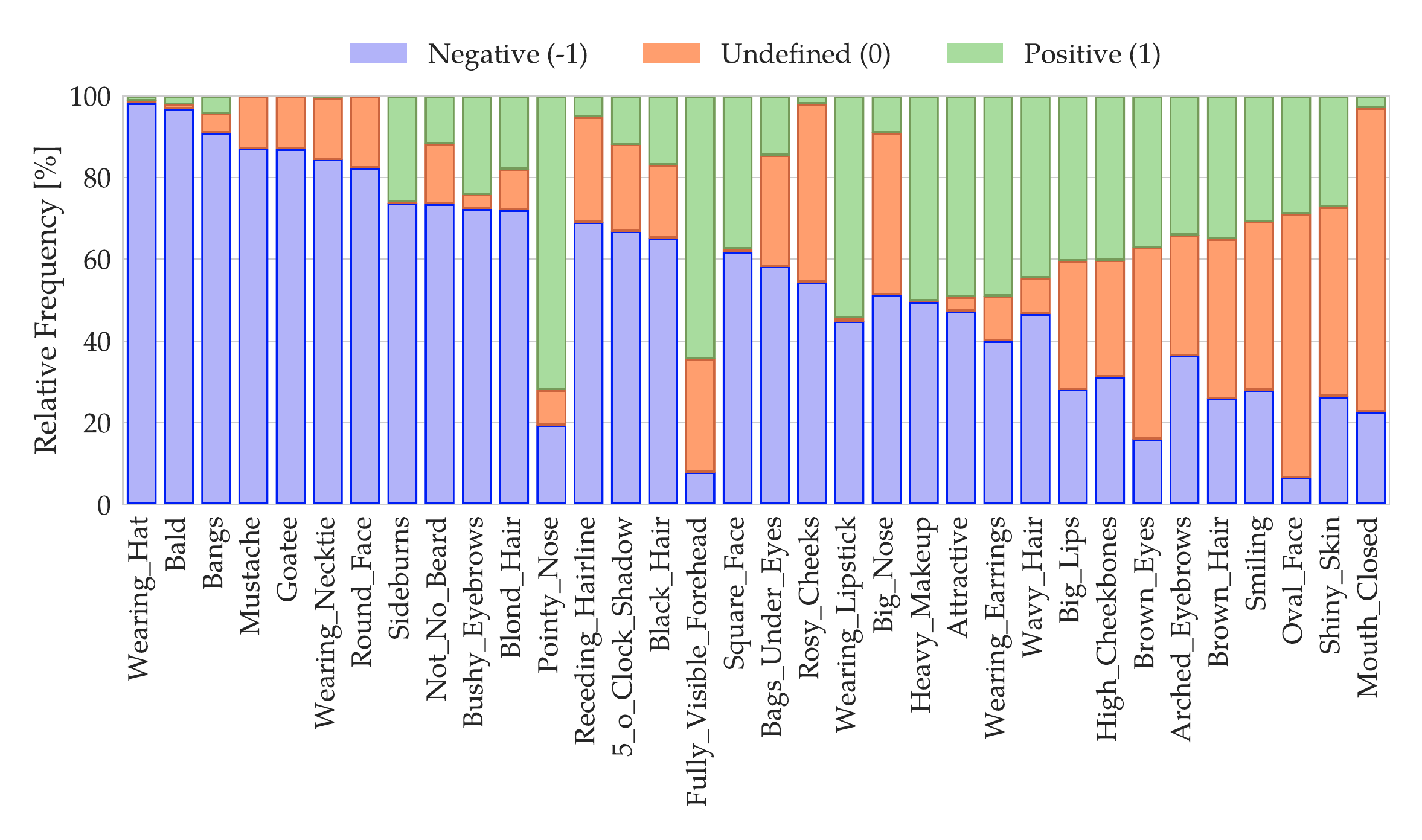}
		\caption{FaceNet}
		\label{fig:assignment-dist-facenet-clustered}
	\end{subfigure}
	\caption{\textbf{Relative frequency of occurrence of attribute-label pairs after filtering for attribute combination achieving highest iGARBE score} - The filter-dictating combination is chosen based on Tables \ref{tab:combined-arcface} and \ref{tab:combined-facenet} for ArcFace and FaceNet, respectively. The shown distributions are computed over those annotated samples in the used database that remain after filtering. Attributes part of the filter-dictating combination are not displayed. The distributions help to understand what gender-fair data might look like.}
        \label{fig:fairness-dist-clustered-single}
\end{figure}

\subsubsection{Combination Correlations}
Previously, we have shown the significant advantage of analyzing combinations of decorrelated attributes to reveal the dictating factors of gender bias in face recognition compared to focusing on individual attributes.
In the following, we scrutinize the single highest-scoring assignment \emph{combination} w.r.t. iGARBE per optimization target model.
We focus on correlations between said combination and other attribute assignments.
In doing so, new insights can be gained since combining decorrelated attributes leads to new correlations that could not have been considered during clustering.
Therefore, the correlations occur based on the assignment combinations, and not directly based on the individual, decorrelated attributes.
Consequently, the spectrum of possible reasons for bias can be tightened even more, solidifying this work's results.

The results of this analysis are shown in \cref{fig:fairness-dist-clustered-single}.
Assignments will be called ``strongly correlated'' if, in the label assignment distribution, a specific label for a given attribute makes up at least \qty{90}{\percent} of total assignments for that specific attribute.
The assignment combination achieving the highest iGARBE score for ArcFace in the ArcFace-optimized analysis consists of negative assignments to \textit{Blond Hair}, \textit{Gray Hair}, \textit{Corpulent}, and \textit{Black Hair} as is presented in \cref{tab:combined-arcface}.
\cref{fig:assignment-dist-arcface-clustered} shows strong correlations of this combination with the attributes \textit{Bald} and \textit{Eyeglasses}, both with negative assignments.
As they are also part of one or more of the remaining top nine assignment combinations, they must carry significance for fairness across genders.
A similar observation can also be made for FaceNet and its respective highest performing assignment combination comprising all negative assignments to \textit{Corpulent}, \textit{Gray Hair}, \textit{Obstructed Forehead} and \textit{Eyewear} as seen in \cref{tab:combined-facenet}.
Said combination strongly correlates with negative assignments to \textit{Wearing Hat}, \textit{Gray Hair}, and \textit{Bangs}.
This is observable in \cref{fig:assignment-dist-facenet-clustered}.
All of the attributes named above are, again, part of the highest iGARBE scoring combinations, underlining their importance for equitable recognition performance.

Overall, the results indicate that the observed performance disparities are strongly influenced by the underlying attribute distributions in the evaluation data.
The findings of this study provide practical guidance for the design of datasets and evaluation protocols in real-world face recognition systems.
\textit{Moving beyond unconstrained demographic comparisons, practitioners can construct evaluation sets in which subjects share comparable appearance attributes, such as hairstyle, facial hair, or accessories, enabling a more unbiased assessment of system performance.} 
For example, balanced evaluation datasets can be constructed such that male and female subjects are matched with respect to selected non-demographic attributes.
This allows performance differences to be analyzed under controlled conditions, reducing confounding effects caused by imbalanced attribute distributions. 
By guiding dataset construction in this manner, future evaluations can yield more reliable and less biased conclusions about system fairness, and enable a clearer distinction between true model bias and artifacts introduced by data composition.

\section{Limitations}
This work relies on an extensive set of facial attributes commonly used in face image analysis. However, a more refined selection of attributes could lead to a more concise description of the factors that causes gender bias. Within the MAAD-Face dataset, there can be assumed to be an unknown small amount of label noise, but we do not believe that this impacts the results of this study.  
This study is further limited to two FRS. Although incorporating additional systems could strengthen the generality of our findings, such an extension is currently infeasible due to the high computational demands of the proposed analytical framework. Despite being designed to efficiently analyse the large number of possible attribute combinations, the experiments still required months of cluster processing time. Given the current constraints in available computational resources, analyses on additional FRS models cannot be conducted at this stage.  
The analysis presented in this work is primarily aimed at revealing the origins of supposed gender bias in face recognition. It may or may not lead to as high-quality results for investigations on age bias or race bias. It should be noted that, while poorly balanced training sets have been ruled out empirically as sources of gender bias, such imbalances in training data may create other problems.  

\section{Conclusion}
As use of FRS becomes ever more prominent in day-to-day activities, ensuring that FRS operate fairly is of utmost importance.  
Understanding the cause(s) of observed differences in accuracy across demographic groups is an essential basis for making judgements about fairness.
Consequently, we conducted comprehensive investigations on the origins of the gender gap in face recognition accuracy, and the role of combinations of decorrelated non-demographic attributes in explaining this gap.
To this end, several novel methodical components were presented and evaluated.
They include \textbf{(1)} a decorrelation-by-clustering toolchain to derive more unbiased statements, \textbf{(2)} two fairness metrics to measure fairness with and without context, and \textbf{(3)} an unsupervised joint investigation framework enabling identification of attribute combinations leading to a vanishing gender gap when adjusting test datasets accordingly.
This notion is reflected in the results with high confidence, leading to the empirical \textbf{observation}:
\begin{boxenv}
	Once male and female subjects share specific non-demographic attributes, the gender gap in recognition accuracy vanishes.
\end{boxenv}
The above holds for all considered experimental setups using two popular face recognition models.
Accordingly, we come to the following \textbf{interpretation}:
\begin{boxenv}
	Gender bias in FRS is likely to originate from non-demographic attributes associated with gender instead of gender itself.
\end{boxenv}
The non-demographic attributes in question relate to the same handful of categories throughout all results.
Specifically, supposed gender bias seems to depend entirely on the presence or absence of specific characteristics related to \textbf{(1)} \emph{hairstyles}, \textbf{(2)} \emph{facial hair}, and \textbf{(3)} \emph{occluding accessories}.
Forcing these attributes to be explicitly shared across gender effectuates a negligibly low error differential between the two subject groups without notably degrading system performance.
Acknowledging these outcomes, which encompass a wide range of reliable empirical investigations, leads to the \textbf{conclusion}:
\begin{boxenv}
	Gender bias in FRS is likely no issue of biology but of the social definition of gender-specific appearance.
\end{boxenv}
Overall, by providing high-confidence explanations, we have performed the groundwork for subsequent evolution of FRS to hopefully overcome this issue.
This work's findings therefore illustrate multiple conceivable routes for future work to explore.
To tackle the gender bias problem in practice, it may be important to focus on making FRS more robust to non-demographic attributes.
If future research wants to further investigate bias mitigation strategies, they need to do so on datasets not only balanced based on gender but also balanced w.r.t. relevant categories of non-demographic attributes.

\printbibliography

\end{document}